\documentclass{llncs}
\pdfoutput=1

\usepackage{graphicx}

\usepackage{natbib}

\usepackage{url}

\usepackage{color}
\usepackage[usenames,dvipsnames,svgnames,table]{xcolor}
\usepackage{latexsym,verbatim}
\usepackage{amsmath}
\usepackage{graphicx}
\usepackage{amssymb}
\usepackage{algorithm}
\usepackage{algorithmic}
\usepackage{authblk}

\newcommand{\out}{\bigcirc}

\newcommand{\simm}{\sim\!\!}

\begin{document}
	
\title{Reasoning in Non-Probabilistic Uncertainty\thanks{Forthcoming with DOI 10.1007/s11023-017-9428-3 in the Special Issue ``Reasoning with Imperfect Information and Knowledge'' of Minds and Machines (2017). The final publication will be available at \url{http://link.springer.com}.}}
\subtitle{Logic Programming and Neural-Symbolic Computing as Examples}

\titlerunning{Reasoning in Non-Probabilistic Uncertainty}

\author{Tarek R. Besold\inst{1} \and
	        Artur d'Avila Garcez\inst{2} \and
	        Keith Stenning\inst{3} \and
	        Leendert van der Torre\inst{4} \and
			 Michiel van Lambalgen\inst{5}
}

\authorrunning{Besold, Garcez, Stenning, van der Torre, and van Lambalgen} 

\institute{Digital Media Lab, Center for Computing and Communication Technologies (TZI), University of Bremen\\
              \email{Tarek.Besold@uni-bremen.de}
           \and
            Dept. of Computer Science, City University London\\
            \email{a.garcez@city.ac.uk}
            \and
            School of Informatics, University of Edinburgh\\ 
            \email{k.stenning@ed.ac.uk}
            \and
            Computer Science and Communication Lab, University of Luxembourg\\ 
            \email{leon.vandertorre@uni.lu}
			\and 
            Department of Philosophy, University of Amsterdam\\
            \email{M.vanLambalgen@uva.nl}
}

\maketitle

\begin{abstract}
This article aims to achieve two goals: to show that probability is not the only way of dealing with uncertainty (and even more, that there are kinds of uncertainty which are for principled reasons not addressable with probabilistic means); and to provide evidence that logic-based methods can well support reasoning with uncertainty. For the latter claim, two paradigmatic examples are presented: Logic Programming with Kleene semantics for modelling reasoning from information in a discourse, to an interpretation of the state of affairs of the intended model, and a neural-symbolic implementation of Input/Output logic for dealing with uncertainty in dynamic normative contexts.
\end{abstract}

\section{Introduction}\label{introduction}
\begin{quote}
	``Almost all everyday inference is \textit{uncertain}, and, thus, human reasoning should be assessed using probability theory, the calculus of uncertainty, rather than logic, the calculus of certainty.'' \cite[p. 308]{oaksford_2004}
\end{quote}

While fully agreeing on the premise of the statement---namely the observation that most human reasoning is uncertain in nature---we want to challenge the conclusion \cite{oaksford_2004} (and many others) draw from it: in our view probability theory is neither the perfect solution solving all challenges introduced by reasoning's inherent uncertainty, nor should logic be overly casually discarded as exclusively fit to deal with reasoning in certainty. In fact,  the conception of monotonic classical logic as `reasoning in certainty' can be misleading.    
In order to substantiate these two claims, below we first illustrate how Logic Programming (LP)---as a  logic-based reasoning and computation paradigm---can be used to model reasoning which involves the resolution of uncertainties of a kind not amenable to probability theory. This is followed by the presentation of a neural-symbolic approach to LP extended to implement  \cite{makinson2000input}'s Input/Output (I/O) logic, which can similarly be used to model non-probabilistic uncertainty in normative reasoning; uncertainty which is compounded by changing or newly added norms (i.e. dynamic environments requiring machine learning).

Among the main foundations of our argument is the observation that there are several qualitatively different kinds of uncertainty, for some of which probability---as convincingly shown by \cite{oaksford_2004}---can be a powerful modelling technique, while others clearly lie outside of the reach of probabilistic approaches. The need for distinguishing kinds of uncertainty has surfaced regularly, for instance, in the study of judgement and decision-making. \cite{knig21:risk} made an early distinction between `risk' (which could be modelled in probability) and `uncertainty' (which could not), in economics. \cite{katv82:variants} already discussed several variants of uncertainty, distinguishing, for example, what we might call private uncertainty (``How sure am I that Rome is south of New York?''  where the truth is taken to be already known by others) from communal uncertainty (``How likely is it at the time of writing  that the euro will collapse?''). 
More recently, \cite[page 1225]{bradley2014types} proposed a taxonomy of uncertainties, in which: \begin{quote} ``[\ldots] ethical, option and state space uncertainty are distinct from state uncertainty, the empirical uncertainty that is typically measured by a probability function on states of the world.''\end{quote} This comes with the claim that a {\em single} probability function cannot provide an adequate simultaneous account.\footnote{However, these authors---somewhat paradoxically---in the end come to the view that whatever uncertainty is the topic, probability is the framework for modelling it; cf. Section~\ref{kindsofuncertainty} for some considerations on the corresponding argument and conclusion.} \cite{mousgig14:risk} expand Knight's distinctions adding a further type of `utter uncertainty' which describes cases where models are used in new domains.

One way of making the existence of a range of kinds of uncertainty distinct from probability generally plausible, is to think in terms of what a probabilistic model has to specify---and then consider what happens when elements are unavailable.  Probabilistic models have to specify an algebra of propositions (variables) which exhaust the `effects' encompassed by the model. Structural relations of dependence and independence then have to be supplied or assumed. Distributional information has to be attached; and last but not least, an assumption of `stationarity' has to be made, i.e. the probabilistic relations between the isolated algebra of propositions have to be assumed to remain constant, and without intrusion of extraneous influences.

Several example phenomena that make this probabilistic schema impossible or unfruitful to apply will recurr throughout this paper: LP modelling of the interpretation of discourse (Section~\ref{logic}); learning (Section~\ref{io2nn}) which contributes various `nonstationarities' to reasoning systems; motivational phenomena which include both the norms expressed in deontic concepts such as priorities, obligations, permissions, contrary to duties (CTDs) (Section~\ref{ANS}), but also motivational ones such as goals and values.  All these various examples share the requirement for the flexible handling of dynamic contexts, with the ensuing robustness to exceptions, as we illustrate.  

Discourses are connected language. The process starts with input of discourse sentences, and sequentially builds a unique minimal `preferred' model at each step.\footnote{Later, we shall consider an alternative LP semantics based on Answer Sets \citep{gelfond}, but we choose Preferred Model Semantics \citep{shoh87:preferred} for now because the uniqueness of preferred models is a crucial feature for cognitive processes such as discourse processing.} The algebra of propositions grows at each step, and when examined carefully, the propositions already in the model change---if only by taking on temporal/causal relations to the new ones that arrived \citep{laha04:prop}---even when they are not dropped nonmonotonically. The structural `common core' of the last LP model in this sequence can provide, for instance, the basis for creating a Bayes Net by adding distributional information \citep{pearl:00,pinosiocommon}. Still, the propositions had not been identifiable until now, and so, no distributions could be attached, nor judgments  made about causal stationarity. Reasoning to this point cannot be probabilistic; and even at this point, it is not entirely clear whether or how the necessary distributional information is available in general. And this is just the beginning of the difficulties since, for example, LP can freely express intentional relations between acts and actors' multiple goals \citep{varga13:teleology}, which can at best be inflexibly `operationalised' in extensional systems. Extensional systems may be able to formulate propositions that, when they become true, indicate that a goal has been fulfilled. However, they cannot capture what goals {\em are}: namely, the abstract flexibility of systems of motivational states and their interactions in unpredictable environments, as we shall see.

It is worth noting that the recent success of probabilistic language models based on neural networks \citep{bordes,graves} is orthogonal to the arguments presented in this paper. Yet, within the area of neural-symbolic computing \citep{garcez_2009}, equivalences have been proved between LP and neural networks, which indicate the possibility of reconciling such apparently very distinct representational frameworks. In fact, later in this paper a (non-probabilistic) neural characterisation of LP will be proposed which seeks to combine the requirements identified here of three-valued semantics and spreading of activation (discussed in what follows) with an ability to learn from examples, which is also a requirement of discourse processing in the case of dynamic environments. This neural-symbolic approach will be applied to deontic attitudes which preclude probability, since they are \emph{motivational} in the extended notion of the word used in this paper: an obligation establishes a goal, even if that goal gets overriden by dynamic changes in circumstance. This will be exemplified in the context of reasoning in uncertainty about an environment of changing norms, which might themselves evolve in many as yet indeterminate ways. 

At this point we take a step back and include what might seem like a mere clarification of terminology, but goes in fact much beyond that: the sense of `extensional' and `intensional' we use here are from the psychological decision making literature---stemming from \cite{tvka83:linda}---in which probability is extensional because its predicates are defined as sets, even though its conditional is non-truthfunctional. In that literature, `intensional' is often synonymous with `informal' as it lacks a suitable logic. Nonetheless, LP (with the semantics as specified by \cite{stvl06:book}) is intensional in the general philosophical sense: its predicates are defined in terms of `senses' which are cashed out as algorithms (completion), and its closed-world reasoning conditionals are `licences for inference' (roughly contentful inference rules) rather than compound propositions (cf. Section~\ref{logic} for an introduction). An example of the role this interpretation of the intensional/extensional distinction plays, is the famous `Conjunction Fallacy' \citep{tvka83:linda} which is supposed to be about judgements of the relative sizes of sets of cases corresponding to predicates and their conjunctions in probability. If a reader's interpretation is in LP, then this extensional distinction makes no sense.  The distinctions required for differentiating qualitatively different kinds of uncertainty can be seen more clearly in the contrast in the semantics of the intensional and extensional systems we discuss below. \cite{kslm15:decision} use the Linda Task (the origin of the supposed Conjunction Fallacy) to illustrate this intensional option for interpretation, and its consequences. In the large, we believe these contrasts between intensional and extensional systems are at the heart of the contrasts in kinds of uncertainty which are our focus.

All these issues concerning the variety of kinds of uncertainty, and the characteristics of different formalisms and representation systems, also resound in the many different types of uncertainty Artificial Intelligence (AI) has to deal with in realistic scenarios: the world might only be partially observable, observation data might be noisy, the actual outcome of actions might be different from the theoretically assumed one (either due to previously unknown or unconsidered factors, or due to independent external influences), or a prediction and assessment of present and future world states, action outcomes, etc. might just practically be outright impossible due to the immense complexity of the environment and scenario.
As can be expected, there is an accordingly large AI literature on reasoning in uncertainty, and the variety of systems other than probability which are available. Quite recently \cite{KernLuka2017} provided a helpful brief map of approaches to uncertainty---which, among others, demarcates logic programming under answer set semantics as distinctively outside both the currently popular system P \citep{kraus1990nonmonotonic} and the AGM axioms \citep{alchourron1985logic}. Another aid to navigation is offered by \cite{halp05:reas} who is concerned with distinguishing different representations of uncertainty, and then studying reasoning from those representations. The example problems he gives in Chapter 1, pages 1 to 4, all have the property that the domain of interpretation and the properties and relations defined thereon are fixed at the outset. This need to fix interpretation is imposed by classical logic and probability, because of their need to generalise about all assignments of values to the vocabulary. Only the sampling, and with it the epistemic states of the judges of uncertainties, varies. For such problems, Halpern argues convincingly that {\em plausibility measures} are a generalisation of probability measures, and that different plausibility measures are more or less appropriate for different problems.\footnote{In terms of concrete examples, the work by \cite{nilsson1986probabilistic} comes to mind as a prominent instance falling within the domain of Halpern's plausibilty measures.} But contrary to this general line of investigation, our focus is on the more radical kind of uncertainty faced by reasoning {\em to} interpretations. 

Our paper thus has a much more modest exploratory goal than Halpern's monumental work. It seeks to develop  two examples in some detail of reasoning {\em to} interpretations. So by definition, in these examples, much less is known about the particularities of their domains until their specification is  finished, and it is the uncertainty {\em during} this process of reasoning which is our focus. A great deal of general knowledge must be applied in the dynamic development of their specification {\em during the period of interpretation whose uncertainties are of interest}. The example of narrative discourse interpretation exemplifies this character:  At the outset of a story, the hearer may know nothing in particular of what will turn out to be in the preferred model of the story that will develop; not even what range of properties and relations will distinguish the characters and events that unfold, even though the hearer's knowledge base must contain a great deal of general knowledge, some of which has to be mobilised to interpret the current input discourse. We nevertheless succeed in reasoning to interpretations of such stories with remarkable, though not inevitable, success. This reasoning is omnipresent in human problem solving and communication \citep{stvl06:book}. It has to be a precursor to modelling in probability, or plausibility measures more generally, because the vocabulary of predicates and relations has to be established by reasoning {\em to} interpretations. Modelling this  reasoning {\em to} interpretations requires a framework that is more radically nonmonotonic than probability theory, whose underlying propositions to which probabilities are attached, are classical logical propositions. \cite{kslm15:decision} argue that such extensional systems are inherently incapable of providing the requisite flexibility. Understanding how reasoning to interpretations works with reasoning from them can lead to a deeper understanding of both. 

Still at this synoptic level, another difference between Halpern's plausibility measures and LP doing discourse interpretation is that the former begin with numerical parameters on propositions, and output numerical parameters on inferred propositions.  In contrast, LP reasoning to interpretations of discourse need have no numerical parameters, even though numerical properties of the {\em logical structures} involved have been shown to be the basis on which reasoners make quantitative judgments of LP conditionals' reliability in inference \citep{kslm15:decision}. Counting the defeaters for a conditional can predict confidence in inferences from that conditional.  For another example, Halpern's conditionals with plausibility measures treat conditionals as propositions having truth values. In LP with Kleene semantics modeling discourse interpretation, conditionals are not propositions (they are licences for inference), and they therefore do not iterate (cf. \cite[p. 184, footnote 9]{stvl06:book}), nor do they become false if they do not apply because of an abnormality condition. 

At a more technical level, there are a number of contrasts between LP (and also the later discussed I/O logic) and plausibility measures \citep{halp05:reas} which adopt the KLM axioms. For example,  plausibility measures apply to systems with  the `OR-rule'.  From $p \rightarrow q, r \rightarrow q$ it follows that $(p \vee r) \rightarrow q$.  This rule does not apply in LP because the `abnormality clauses' defeat it: $p \wedge \neg ab' \rightarrow q, r \wedge \neg ab'' \rightarrow q$ it does not follow that $(p \vee r) \rightarrow q$. 

We close this differentiation of our focus from existing work with an example of  a more empirical dissatisfaction with probability as a model of human reasoning.  \cite{tvka74:hospital} and \cite{abc99:simple,giger11:adaptive} have extensively developed evidence that human reasoning is heuristic with regard to probability models.  In an especially clear case of this relation, \cite{jusnilss09:not-prob} argue that peoples' judgements of likelihoods in uncertainty do not obey probabilistic models.  However, their concern is with heuristic approximations for combinations of probabilities.  They assume that people have access to the probabilities (or their estimates), and produce evidence that people combine them in the conjunctive case by `weighting and adding', rather than the probabilistically normative procedure of multiplying.   This is a different, calculative level of issue with probability than the conceptual differences that concern us here. These authors tiptoe towards the cliff of appreciation  that probability might not be the right normative theory:
\begin{quote} ``Importantly in this context, to the extent that problems are framed in terms of probability that often requires multiplicative information integration, the strong inclination for linear additive integration is a first sense in which probability theory may often not be a very useful (or useable) guide in life.'' \cite[p. 861]{jusnilss09:not-prob}\end{quote}
Nonetheless, they do not take the leap of offering an intensional framework.
\cite{kslm15:decision} explore an incorporation of such heuristic approaches within LP to provide a simple probability-free model of judgement and decision.  This approach offers insights into how intensional reasoning produces structural foundations for  subsequent probabilistic modelling through their `common core' \citep{pinosiocommon}.  This functional relation of the two contrasting types of reasoning in uncertainty is of central importance in human reasoning, and is possibly one of its main ingredients which is least represented in current AI.

In Section~\ref{logic} we first focus on the application of LP in building process models of ``reasoning {\em to} an interpretation'' as a crucial part of human reasoning. As, for instance, \cite{stvl08:oaks} argue, when faced with interpretational uncertainty about the information involved in a problem, we cannot engage the  computational complexity of probability, but have to take more accessible inferential paths---which are often quite successful. Here, among other work building on the book by \cite{stvl06:book}, a version of LP, using a semantics based on three-valued Kleene logics, offers itself as a modelling approach. Since it is a crucial cognitive capacity involved in our daily lives, as already mentioned above, discourse processing will serve as core theme and paradigmatic example in this section.

Section~\ref{neural-symbolic} subsequently takes a  more AI-centric view and describes a neural-symbolic architecture combining the I/O logic proposed by \cite{makinson2000input} 
with artificial neural networks (ANNs), applied to normative reasoning tasks involving uncertainty introduced by changing norms over time. I/O logic seeks to analyse the sometimes subtle asymmetries between what can be fed into, and what can be output from, other logics, wherever there are inputs that cannot occur as outputs, or {\em vice versa}.  Normative reasoning is chosen because deontic systems have to encode obligations and permissions which will be manipulated as propositions  by the `inner logic', but whose force is not  entirely captured by such manipulation. 
For example, consider `cottage regulations' often discussed in deontic logic. When the input says that there is a dog, and the output says that there should be no dog, then there is a violation (which could be sanctioned). Alternatively, if the input says that there is no dog, and the output says again that there should be no dog, then there is a fulfilled obligation (which could be rewarded). Moreover, we may also have according-to-duty relations such that the output says that there should be no fence either when the input says there is no dog, or when the input is just a tautology; and contrary-to-duty (CTD) relations in which case the output may say that it is obligatory that there is a fence when the input says that there is a dog. Examples like these will be discussed later. Simply put, the challenge of norm dynamics is to change such---already by themselves quite complex and often highly interdependent---relations between input and output when new information becomes available that norms have changed. Interestingly, in neural networks, as in I/O logic, what can be fed into and what can be output from a neural network is strictly defined, in contrast to LP. This will be analyzed in detail. The neural part of the neural-symbolic approach enables the required form of learning, and learning introduces a dynamics to normative systems that also exercises I/O logic, introducing more kinds of uncertainty differentiated from probability.

Section~\ref{conclusion} then concludes our argument and sketches several directions for future work.

\section{Logic Programming Modelling Reasoning to an Interpretation}\label{logic}

In this section we focus on the applicability and advantages of an LP-based approach for modelling reasoning in interpretatively uncertain situations, i.e. situations in which there is uncertainty concerning the relevance or precise meaning/interpretation of propositions. Section~\ref{LPmodelling} introduces the relevant form of LP using three-valued Kleene semantics, and conceptually motivates its use as a modelling tool for this human reasoning. This is followed by a more focused treatment of this type of LP applied to dealing especially with the uncertain aspects of reasoning in Section~\ref{kindsofuncertainty}.

\subsection{Logic Programming Modelling Reasoning}\label{LPmodelling}

LP is a formal system that has been used extensively to model reasoning in a variety of domains, as widely separated as motor control \citep{Shanahan:02}, and imitative learning \citep{varga13:teleology}. Its employment in cognitive modelling grew primarily out of an analysis of discourse processing, in particular of interpretation \citep{laha04:prop,stvl06:book,vast16:hndbk}. People take in sentences online, in fractions of a second,  and effortlessly update their current  discourse  model, fully indexed for co-references of things, times and events, and their temporal and causal relations. This may require  far-flung bits of  background knowledge, retrieved from a huge knowledge base (KB) of semantic memory composed of conditional rules understood as licences for inference. Discourse interpretation involves constructing a preferred or intended model\footnote{\label{intended}{\em Intended model} is the psychological notion which corresponds to {\em minimal model}. {\em Model} is to be read as semantic model. {\em Preferred model} is the logical notion used by \cite{shoh87:preferred}. Keeping in mind the terms belonging to different fields, we use intended, minimal and preferred as synonyms.} of the context. A crucial application of LP is modelling the efficient inferential retrieval from long-term memory of the relevant cues needed to construct or decide on an interpretation of the state of affairs as basis for, e.g., choosing a course of action.

Following \cite{stvl06:book},  we view reasoning to be a two-stage process. It starts with reasoning {\em to} an interpretation (the computation or retrieval of a meaningful  model of the current situation), which may be followed by reasoning {\em from} the interpretation. In a typical (i.e. cooperative) conversational context this amounts to computing the model the speaker must have intended to convey. The computation of the intended or preferred model by LP complies with cooperative Gricean principles, but is much more efficient than computing directly with those principles. Imagine you are talking to a friend who is telling you her holiday stories, including an outing into the countryside by car. She says: ``And then I press the brake! And then $\ldots$''. LP is constructed so that it interprets this utterance online under the assumption that the speaker's goal is to provide  everything the hearer needs to know and nothing more, in order to construct a minimal model of the discourse at each point.  At this point that model predicts  a continuation of the story consistent with the car slowing down, though this may well be retracted at the next point. It does not consider, without positive evidence, that there might have been ice on the road, or that hers functions differently from all other cars, (or any other defeater). LP with negation-as-failure and Kleene's strong semantics  has shown to be a good candidate for a suitable logic of such cooperative intensional  reasoning.  

LP conditionals function as  `licenses for inference', rather than  sentences compounded by the `if \ldots then' connective. They can be thought of as highly content-specialised rules of inference which are always applicable when a clause matches them. But it has the important consequence, already mentioned, that---like natural language conditionals---LP conditionals do not iterate, especially in the antecedent, to produce compound propositions, and do not themselves have truth-values: they are assumed applicable on an occasion unless evidence of exception arises. 
LP as discussed by \cite{stvl06:book} and \cite{laha04:prop} formalises a kind of reasoning which uses closed-world assumptions (CWAs) in order to keep the scope of reasoning to manageable dimensions by entities with limited time, storage, and computational resources, though with very large knowledge bases (KBs)---such as we are. Historically, LP is a computational logic designed and developed for automated planning \citep{Kowa88:logicprog,doet94:logi} which is intrinsically preoccupied with relevance---making for an important qualitative difference between LP and classical logic (CL). 
Returning to the type of communicative situation as just introduced in the car example, discourse processing in LP is cyclical. When a new input sentence arrives, its terms are searched for in the KB. The existing minimal model of the discourse to this point is then updated with any new relevant information, according to CWAs, and the cycle repeats. CL, in contrast, assesses the validity of inferences from premises to conclusions with respect to  {\em all possible models} of the premises, so the question of relevance does not even arise in CL, except outside the logic in the framing of problems. So LP is  {\em not} just a poor man's  way of doing what can be done better but with more difficulty in CL---the two different types of logic simply serve different incompatible reasoning purposes, and in this sense are incommensurable.

The basic format of CWAs is the one for reasoning about abnormalities (CWA$_{ab}$), which prescribes that, if there is no positive information that a certain abnormality-event must occur, it is assumed not to
occur.\footnote{{\em Abnormality} is a technical term for exceptions and should not be taken as having any other overtones. Some terminology: in the conditional $p \wedge \neg ab \rightarrow q$, $ab$ is the schematic abnormality clause. A distinct $ab$ is indexed to each conditional, and stands for a disjunction of a list of defeaters for that conditional; CWA$_{ab}$ is the CWA applied to abnormality clauses; $\neg$ is the 3-valued Kleene connective, whereas  negation-as-failure is an inference pattern that results in negative conclusions by CWA reasoning from absence of positive evidence; falsum ($\bot$) and verum ($\top$) are proposition symbols which always take the values false or true respectively; turnstile ($\vdash$) and semantic turnstile ($\models$) are symbols indicating syntactic and semantic consequence respectively.} 
These  abnormalities are with respect to the regularity expressed by a conditional; for example, ice on the road causing a car not to stop although the brake is pressed, is an abnormality with respect to the default functioning of brakes. Potential abnormalities are included in the LP meaning of the conditional $\rightarrow$, hence what is labeled `counterexample' in CL does not invalidate a conditional inference in LP, it is merely treated as an exception. The conditional has an operational semantics as an operator  that modifies truth values of atomic formulas. This is the logical mark of its use as an exception-tolerant  `licence for inference'. It is represented as  $p \wedge \neg{ab} \rightarrow q$; it reads as `If $p$ and {\em nothing abnormal is the case}, then $q$'. The antecedent $p$ is called the {\em body} of the clause, and the consequent $q$ is its {\em head}.\footnote{Another remark concerning terminology: while in this context the use of the terms `head' and `body' is commonplace in computer science, we will in the following restrict ourselves to `antecedent' and `consequent' in order to maintain homogeneity also with terminology in philosophy and logic.} $p$ and $q$ are composed of atomic formulas, with $q$ restricted to literals (atomic formulas or their negations) and $p$ to conjunctions of literals. The only connectives that may occur in the antecedent are $\wedge$ and $\neg$.  

Sets of such conditional clauses constitute a general logic program $\mathcal{P}$ (i.e. a program allowing negation in the antecedents of clauses). $\mathcal{P}$ can be understood as a recipe for computing a unique model of the discourse it represents, on the basis of information from background knowledge inferentially selected as relevant. Negation in the antecedent requires utilisation of a three-valued semantics.\footnote{Cf. \cite[chapter 2]{stvl06:book} for the justification, or the Appendix in \citep{kslm15:decision} for a more succinct version.} We opt for strong Kleene semantics for LP, where the middle truth value {\em u}  means `currently  indeterminate'; it is not a graded truth akin to a probability---as in Lukasiewicz three-valued logic---but rather a stage of computation of an algorithm which can evolve towards either {\em 0} or {\em 1}. 
This gives LP the sort of semantic mobility which constitutes  a crucial reason for claiming that it can capture the particular kind of nonmonotonic flexibility of reasoning required to deal efficiently with interpretational uncertainty (cf. Section~\ref{kindsofuncertainty}). Facts $q$  can be represented in logic programs as consequents of tautologies, $ \top \rightarrow q$ (for simplicity we write only $q$). The CWA$_{ab}$ requires that for an initial interpretation at least, abnormalities are left `at the back of the reasoners' minds', i.e. outside of the {\em minimal  model} of $\mathcal{P}$. However conditional clauses have conjoined abnormality conditions of the kind $r_1 \rightarrow ab_1, \ldots r_n \rightarrow ab_n$; when evidence of $r_1, \ldots r_n$ becomes available, it activates the correspondent $ab_k$. If we take $\mathcal{P}$ = \{If the brake is pressed, the car will slow down\}, the fact `there is ice on the road' is not represented in its minimal model because this model  disregards all potential abnormalities without explicit evidence. The  information  in a minimal model  includes all that is currently known to be relevant and nothing more than that; the only relevant information is explicitly mentioned or derivable.  The derivation capacities of LP's search of its KB are what achieves this computation of relevance. If new input is that there was a storm last night, then this, together with other information in the KB about local meteorology, may yield an inference that there is ice on the road. The input does not mention ice, but ice on the road is a relevant defeater for braking causing slowing, that we assume is already in the list of defeaters for the brakes conditional.  Of course, the new information will lead to many other potential inferences during the sweep through the KB, but if they do not connect with anything in the current model, these inferences will not be added because there is no evidence of their relevance.  Sometimes such retrieval can involve several conditional links.  This is  a computationally explicit example of what psychology knows as `spreading activation' models of memory retrieval, but has been neurally implemented for `propositional LP' as described in \cite[chapter 8]{stvl06:book}.  Spreading of activation will also be given a slightly different  implementation in the neural networks used later in this paper. Such networks will adopt an answer set semantics \citep{gelfond} allowing the use of \emph{default} negation for implementing CWA explicitly, and \emph{classical} (sometimes called explicit) negation, which allows for reasoning as intended here, that is, in the absence of a proof of A and its negation, the truth-value of A is `currently  indeterminate'. 

LP models embody much information about stereotypes.   For a micro example, it is part of the stereotype of our braking scenario, that the car that is  conjured is in motion when the brake is pressed, as developed below. Minimal models are at the intersection between the current input (be that heard language, or observations of any other form) and the reasoner's background knowledge; they contain all the relevant information and nothing more. In logical terms, they are constructed using the two-step rule of {\em completion} ($comp$): 
\begin{enumerate}
	\item take the disjunction $\vee$ of all antecedents $p_1, \ldots p_n$ with the same consequent $q$ in program $\mathcal{P}$;
	\item replace $\rightarrow$ with $\leftrightarrow$.\footnote{Here, $\leftrightarrow$ denotes a classical biconditional in the object language.}
\end{enumerate}

Thus, if $\mathcal{P} = \{p_1 \rightarrow q, \ldots p_n \rightarrow q\}$, $comp(\mathcal{P}) = \{p_1 \vee \ldots p_n \leftrightarrow q\}$. The minimal model of $\mathcal{P}$ is in fact a model of $comp(\mathcal{P})$. Such a model does not include information that is either not explicitly mentioned in $\mathcal{P}$, or not derivable from it, and in this sense it is minimal. The CWA$_{ab}$ provides the notion of valid inference in LP---an inference is valid if it is truth-preserving with respect to the minimal model of the premises. The contrast with CL is noteworthy: in CL an inference is valid if and only if the conclusion is true in all possible models of the premises. Because of this LP turns out much less computationally complex than CL, and therefore is a promising candidate framework for realistic performance models of fast, implicit, or automatic reasoning (online discourse interpretation being the paradigmatic case).\footnote{Basically, in LP queries can be answered in time linear with the length of the shortest inferential path in the KB.}

Once a minimal model is constructed, further reasoning {\em from} that model ensues according to the derivation rule of {\em resolution}, or the reduction of goals to subgoals by means of backwards chaining.\footnote{This direction of reasoning is from effect to cause, from goals to subgoals, or simply backwards in time. The typical backward inferences  are modus tollens ($p \rightarrow q, \neg q \models \neg p$) and affirmation of the consequent ($p \rightarrow q,  q \models p$), initiated by the consequent. The psychological findings that these inferences are more difficult might well be a result of the micro-scale of the tasks being used \citep{slolag15:causality}, or of the slightly more complex form of the CWA$_{ab}$ needed \cite[pp. 176--177]{stvl06:book}.} The syntactic manifestation of the CWA$_{ab}$ is the inferential rule of negation-as-failure, which is applied restrictively to proposition letters which are consequents of falsum, i.e. to $q$ occurring in formulae like $\bot \rightarrow q$. {\em Queries} or {\em goal clauses} initiate derivations. \begin{quote} ``Operationally, one should think of a query $?q$ as the assumption of the formula $\neg q$, the first step in proving $q$ from $\mathcal{P}$ using a reductio ad absurdum argument. In other words, one tries to show that $P,\neg q\models \bot$. [\ldots] one rule suffices for this, a rule which reduces a goal to subgoals.'' \cite[p.230]{laha04:prop}\end{quote}

Suppose one wants to reduce the query $?q$ with respect to the program $\mathcal{P} = \{p \wedge \neg ab \rightarrow q, \bot \rightarrow ab\}$, which contains no positive information about abnormality. Negation-as-failure allows the reduction of $q$ to $p$, because $ab$ is a consequence of falsum. 

The formal parameters of LP, e.g., the semantics based on truth in unique minimal models, the nonmonotonic definition of validity, or the syntactic rules, such as negation-as-failure, recommend it for applications in the modelling of human reasoning. In the first place, individuals' background KBs are sets of exception-tolerant regularities. Further, LP provides a `direct route' to (the reasoner's beliefs about) states of the environment which ground inferences: observations can be looked at as the effects of those (beliefs about) states, represented as literals occurring in the consequents of  KB clauses.  We saw above that the LP syntactic rule of resolution amounts to stepwise backward reasoning from goals (effects) to subgoals (causes), i.e. from queries to their causes  represented in the bodies of  conditionals. This matches conceptually with the plan-like psychological structure of reasoning strategies: starting from the goal, i.e. a particular desired state of affairs that  calls for action, one attempts to derive a behaviour that presumably leads to that state. Similarly, when a certain state is observed, backwards reasoning derives as its cause the (beliefs about) states which appear in the antecedent of the clause whose consequent it is. This `effect to cause'  inference takes place with respect to the minimal model of  $\mathcal{P}$, after the operation of completion has been performed; further, it is made under the auspices of $CWA_{ab}$ in the syntactic form of negation-as-failure. Therefore the (beliefs about) states are presumably, to the best of the reasoner's knowledge, relevant for the current inference---they are the most plausible conditions for the observed effect to occur. 

Tying back into our overarching theme for this article, it should be noted that LP has some close correspondences with probabilistic models. At least in causal cases, LP produces the structural features of the models which are shared with the structural features of causal Bayes Nets \citep{pearl:00,pinosiocommon}. Nevertheless, the computational properties are highly distinct \citep{bgstvl13:hndbk,stvl08:oaks}. Probability cannot be the basis for what LP does in discourse processing. Discourse processing at a semantic level requires extremely fast reasoning about novel arrangements of properties and goals, currently unknown by the reasoner to be relevant to the discourse,  in order to identify the propositions underlying the discourse, and this without knowledge of the distributions required for probability models.  Probability has the general computational feature that the heavy computation of critical probabilities has to be done when the information defining the problem is complete (i.e. the relevant propositions identified and refined), and this is not viable in on-line discourse processing. Foreshadowing the discussion in the following subsection, the relevant type of uncertainty here is uncertainty of interpretation, not uncertainty about truth values or their probabilities. We strongly suggest that this state of affairs is not limited to discourse processing, but continues into many of the situations where people must interpret novel information. 

\subsection{The Many Logical Faces of Uncertainty, or What Logic Programming Can Do for Reasoning}\label{kindsofuncertainty}

We have introduced LP modelling of discourse interpretation in some detail because it is perhaps the extreme example of reasoning to interpretation. Once this contrast with probability models is established, it enables us to understand its kind of uncertainty as qualitatively distinct from probability. For this we resort to comparison of the logical systems that define them.  In contrast to most of the proposals of kinds of uncertainty listed in the introduction, we see the necessary focus as on differences in the epistemic relations of the  user of a logic to the propositions expressed, rather than in the content of the propositions themselves.  To illustrate with examples of the version of LP just described, a speaker and a hearer are modelled as cooperating for the speaker to communicate to the hearer her intended preferred model by uttering a sequence of sentences. As is typical in cooperative action, this is a non-zero sum game.  If the hearer doesn't get the model, then the pair have failed (blame is another matter irrelevant here).   Once  this cooperative nature is captured in the semantics, it becomes evident that there is a concomitant kind of necessity: one might call it communicative necessity.  If the speaker says "Once upon a time there was a cat and a dog" the hearer can conclude with certainty that the intended model contains two distinct animals {\em at this point}.  This is not certainty about the real world (whatever that is).  It is certainty about the expressed intended model, where the relation of that model to the world is for the time being suspended.  However complex the gyrations in the continuation are that perhaps reveal that the cat was actually  mistaken for the very dog, so that the model communicated then has only one animal in it,  the initial proposition that there are two animals is `communicatively necessary' in this logic of interpretation.  What is certain is the interpretative facts (as long as the discourse is clear, which in turn depends whether the speaker and hearer's KBs are well aligned in their relevant parts):  not  any existing situation in the world.  Fictional examples are extreme and therefore helpful, but in fact much of our communication starts out nearer to this kind of process than to the construction of a state of affairs that the hearer already knows how to anchor to the world. With `true' stories, we also do not know   what is going to happen, or even who is going to turn up, until they are done.  Once the appropriate kind of necessity is grasped, it is easier to see the kind of uncertainty that is involved.  The hearer is uncertain about what information will arrive, and what updating of the model will therefore be required to form the propositions that will be in the final model.

Different disciplines over the last half century or so have contributed to a greater understanding of the range of logical systems for treating uncertainty. To see the importance of a qualitative classification of kinds of uncertainty, consider CL as distinct logic also exhibiting a characteristic kind of uncertainty while reasoning toward a proof of a conjecture.\footnote{Nota bene: This stands in harsh contrast to \cite{oaksford_2004}'s above quoted conventional characterisation of CL as `reasoning in certainty'. When this way of conceptualising monotonic CL in contrast to nonmonotonic logics was introduced in the mid-1970s and 1980s---for instance, in the wake of \cite{minsky}'s frames or \cite{mccarthy}'s circumscription approach--- the underlying concern was not with characterising kinds of uncertainty, but with contrasting two systems on the one specific property of (non)monotonicity. } This is reasoning in uncertainty about whether the  conjecture is a theorem, and it cannot  be measured by probability. A proof dispels this uncertainty with a positive answer, while a counterexample resolves it with a negative answer. For another example, deontic logic defines reasoning in moral uncertainty toward moral necessity. Each logic has its own kind of uncertainty. If there were no uncertainty, the entire motivation behind  reasoning would be more than questionable. Also---coming back to the remark concerning the kinds of uncertainty treated by a logic being twinned with the logic's kind of necessity---the CL example can serve as prime example: CL's kind of necessity is truth in all models of the premisses, which is distinctively {\em not} explicable in terms of probability.

First, a general characterisation of the problem of distinguishing kinds of uncertainty, before returning to the example.  Any kind of uncertainty is a three-way relation between a person, their epistemic state, and  a proposition.  A probability model is one kind of specification of an epistemic state, which assigns probabilities to component propositions.  This is not a point about the subjectivity or otherwise of probability.  However objective the probability of an event may or may not be, it also depends on the epistemic state of the assigner---what they know or believe that is expressed in the relevant probability model.  This epistemic state is also objective.  Epistemic agents with different knowledge and belief about the same objective events will generally rationally assign them different probabilities on the basis of different models. Think of the players in a card game who know their own but only their own hand: each has a different probability model assigning different probabilities, to say, the next card played. 

If the epistemic state of an agent does not permit them to specify a probability model by identifying the  propositions relevant to their epistemic state, then their uncertainty will be of a different kind than probability. The authors cited in Section~\ref{introduction} as advocating varieties of uncertainty---with the exception of \cite{mousgig14:risk}--- look to identify them through probability models. For example, take the most elaborated classification by \cite{bradley2014types}. It starts out promisingly qualitative, as demonstrated by the quote above. But the authors still end up with probabilities. The authors' argument is only that they cannot be assigned a {\em single} probability function. But they do not  consider the variety of logics that are required to characterise the epistemic states that constitute the uncertainties. It is the characterisation of epistemic states that requires the different logics; not particularly  the nature of the `output proposition' to which the uncertainty is assigned. 

Again using discourse processing as example, LP describes the sequence of epistemic states the hearer goes through as they interpret a discourse. In this view, LP---as candidate for a logic underlying cooperative discourse semantics---specifies a process which  takes in new information about the current context of reasoning, and interprets it in the light of background KB of regularities into a unique minimal model which identifies the relevant propositions. Consider the case of a discourse-initial: ``Max fell. John pushed him''. First it is  crucial to see that the discourse model we construct is more than the set of  sentences. The two sentences ``Max fell'' and ``John pushed him'' are logically unrelated, but the  model we derive (perhaps unwittingly) is extensively augmented with new information.  For example, the link attributed is causal---Max's falling follows John's pushing in time,  as its effect---and  there is  a spatial contact  between Max and John. And the model specifies that Max is not John. And so on.  This extra material is derived based on our general knowledge about human-on-human pushings and fallings. The example is chosen because here the KB actually induces an interpretation where the sequence of events departs from the sequence of narration, making the reasoning more prominent. 
Of course, it quickly becomes obvious that here context is everything. If Max and John were fish in water, we would struggle to interpret, because falling in water is hard for a fish, and not the typical result of pushing. If, on the other hand, the next clause were to be a continuation of the second sentence: ``[\ldots], or what was left of him after hitting the ground, over the cliff.'', we would have to revise the model in the light of the new information. Now, the pushing is subsequent to the falling (whose causation is unknown), and there is a cliff in the model, near to the protagonists. The uncertainty resides in the fact that we do not know what information is going to be involved; and this information is not available until we have integrated the current input and our background knowledge into the developing model, resolving temporal, causal and referential relations, among others. We will not present an LP formalisation here (cf. \cite[p. 131]{laha04:prop}): our purpose is to point to the omnipresence of such inference, and the richness of the general knowledge that has to be found and applied, generally at speed, and without awareness.  Electrophysiological observations on analogous materials strongly corroborate these inferences' occurrence \citep{pivl10:autism}. 

What is also important is that---as already stated above---interpretative inferences define their own kind of necessity, namely communicative necessity. They entail nothing about how the world has to be, but they do entail what we have to take as the speaker's intended model of her discourse. This does not mean we have to take it as truth: we may even have good reason to believe it is part of a deliberate deception. But if we need to understand that deception, we must first understand the intended model to see what pack of lies is being offered, i.e.  what has been communicated. The general point is that we {\em must}  construct our interlocutors' intended models if we are to communicate.

But now, returning to our overarching question, how does this relate to the types of uncertainty probability theory can and cannot cover? As we engage in this little cliff top drama, we are uncertain at every new discourse addition what meanings are involved. Until each new sentence has been successfully incorporated into the model we do not know which sense it expresses. But the arrival of new propositions is just the tip of the iceberg. We also do not know its effect on the other bits of meanings that were there before. They may have disappeared completely, and they probably have changed, if only by the relations that are now determined between the old and the new. In contrast, a probability model has to be founded on a set of propositions where such relations are explicit. So at best, our discourse could invoke a sequence of probability models, one at each step. But there need not be any systematic relation between one and the next. The only one likely to be of much interest as the foundation of a probability model is the `final' one (cf. \cite{stvl08:oaks,bgstvl13:hndbk}). And if one wanted a probability model of this last one, then the corresponding LP model provides exactly the common structural core on which the necessary probability information would have to be hung, perhaps even provided by estimates from the conditional frequencies available in the LP net involved (cf. \cite{kslm15:decision}). In other words, while probability theory does not deal with the dynamics of uncertainty about interpretation, LP can serve as a modelling approach for this crucial part of human reasoning (remember  the 3-valued Kleene semantics we use for LP, in which the $u$ truth value can evolve during computations towards $1$ or $0$).

So discourse processing provides our first example of a prevalent cognitive process which deals in uncertainties and certainties of a kind not treated by probability. And LP provides an alternative logic for this examples' analysis.   

\section{Neural-Symbolic Computing Modelling Dynamic Normative Contexts}\label{neural-symbolic}

Following our presentation of LP as a promising approach to modelling reasoning to an interpretation (and resolving the associated communicative uncertainty), in this section---building upon and expanding the ideas from Section~\ref{logic} conceptually as well as formally---we focus on normative reasoning and the associated form of uncertainty resulting from dynamic changes or expansions of norms. Regarding aspects of uncertainty, norms and norm-based reasoning pose several challenges: among others they tend to be highly sensitive to the context the reasoner finds herself in and her interpretation thereof (e.g., when deciding which norms apply, or---if several options could be chosen from---which would be preferred options, either due to the resulting actions or to abstract value-related considerations), and usually are subject to change over time (e.g., when existing norms are altered in content or interpretation, or new norms are introduced). These properties establish a natural connection to the virtues of LP-based models of reasoning described above: LP's construction of preferred models can be seen as construction of contexts. Still, while in the previous section we were mostly focused on the application of LP in modelling human reasoning, we now shift emphasis to a more AI-based perspective, considering reasoning in intelligent agents in general.
In artificial social systems, norms serve as mechanisms to effectively deal with coordination in multi-agent systems (MAS). Among the open problems relating to the use of norms in these systems is how to equip agents to deal effectively with norms that change over time \citep{Boella-etal:09normchange}, either due to the introduction of new norms, due to explicit changes made by legislators to already existing norms, or due to different interpretations of the law by judges, referees, and other judicial bodies.\footnote{Terminology yet again: for the purpose of this article we use law, norm, rule, etc. as synonymous.} 

In trying to tackle the difficulties arising from the dynamic nature of norms, we combine I/O logic \citep{makinson2000input,makinson:ciol,makinson:perm} with neural-symbolic computation \citep{garcez_book} in order to propose a formal framework for reasoning and learning about norms in a dynamic environment. I/O logic is a symbolic formalism---in several ways closely related to LP as will become apparent below---used to represent and reason about norms, providing reasoning mechanisms to produce outputs from the inputs, each of them bearing a specific set of features. The neural-symbolic paradigm of \cite{garcez_book} on the other hand embeds symbolic logic, and in particular LP into ANNs. Neural-symbolic systems provide translation algorithms from symbolic logic to ANNs and vice-versa: the resulting network is used for robust learning and efficient computation within a connectionist framework, while the logic provides background knowledge to help learning, as the logic is translated into the ANN, and high-level explanations for the network models, when the trained ANN is translated into logic. 
The combination of logic and networks is achieved by representing the I/O logic within the computational model of ANNs, leveraging a similarity between I/O logic and ANNs: both have separate specifications of inputs and outputs. We exploit this analogy to encode symbolic knowledge expressed as I/O logic rules into a standard ANN, and use the resulting ANN to learn new norms in a dynamic environment. Thus, two main steps have to be achieved, namely the translation of I/O logic rules into ANNs, and the evaluation of the ANN learning mechanism at refining normative rules in time.

With the exception of game-theoretic approaches (cf., e.g., \cite{SenA07,BoellaT06,ShohamT97}), few machine learning techniques have been applied to tackle open problems like revising and learning new norms in open and dynamic environments. We show how to use ANNs to cope with some of the underpinnings of normative reasoning---namely \emph{permissions}, \emph{CTDs} and \emph{exceptions}---by using the concept of priorities between I/O (or LP) rules, i.e. LP rules with metalevel priorities \citep{Antoniou98}. Thus, the contribution here is in allowing the handling of the uncertainty associated with norm changes by combining symbolic and sub-symbolic representations to provide a flexible and effective methodology for learning, normative reasoning, and specification in MAS.
After a short introduction to neural-symbolic integration and the corresponding conceptual and architectural paradigm in Section~\ref{nss}, I/O logic is formally introduced in Section~\ref{i/o}, explaining abstract normative systems, propositional I/O logic, and the notion of permissions in the corresponding normative framework. Section~\ref{architecture} then gives an overview of the neural-symbolic architecture implementing I/O logic, before Section~\ref{normativepriorities} shows how priorities can be used to encode and regulate certain types of normative problems.
Section~\ref{ncil} then finally draws all pieces together in presenting the resulting system for normative connectionist learning and LP.

\subsection{Neural-Symbolic Systems}\label{nss}

The main purpose of \emph{neural-symbolic integration} is to bridge the gap between symbolic and sub-symbolic representations. To this end, neural-symbolic systems bring together connectionist networks and symbolic knowledge representation and reasoning \citep{garcez_2015}. In this way, neural-symbolic systems seek to take advantage of the strengths of each approach whilst hopefully avoiding their drawbacks. For our current purposes, we are particularly interested in three consecutive steps: representing the norms governing a normative system formally and soundly in an ANN, using the network to achieve efficient parallel computation, and finally exploiting the instance learning capacities of ANNs to adapt the norms in the system through learning. This should give rise to a normative system capable of integrating reasoning and learning capacities in an effective way. In what follows, we introduce the basic concepts of ANNs and neural-symbolic systems used in this article, with an emphasis on an extension of the \emph{connectionist inductive learning and logic programming} (CILP) system by \cite{garcez_book}.

\begin{figure}[htb]
	\centering
	\includegraphics*[height=4.5cm]{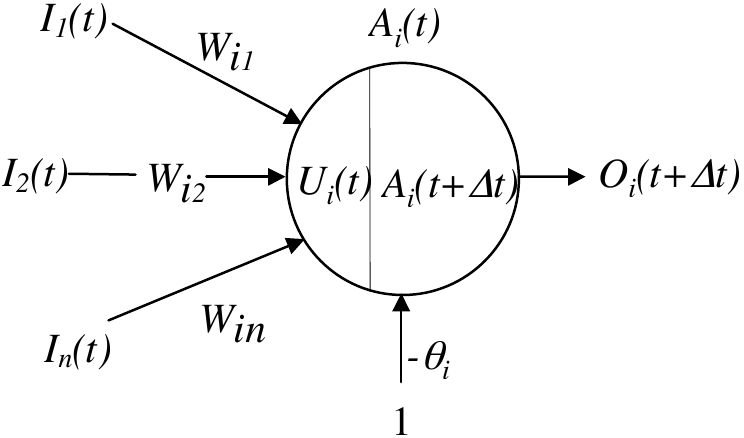}
	\caption{The neurone or processing unit.}
	\label{neura}
\end{figure}

An ANN is a directed graph with the following
structure: a unit (or neurone) in the graph is characterised, at time $t,$ by
its \emph{input vector} $I_{i}(t)$, its \emph{input potential} $U_{i}(t)$,
its \emph{activation state} $A_{i}(t)$, and its \emph{output }$O_{i}(t)$.
The units of the network are interconnected via a set of directed and
weighted connections such that if there is a connection from unit $i$ to
unit $j$ then $W_{ji}\in \mathbb{R}$ denotes the \emph{weight} 
of this connection. The input potential of
neurone $i$ at time $t$ ($U_{i}(t)$) is obtained by computing a weighted sum
for neurone $i$ such that $U_{i}(t)=\sum_{j}W_{ij}I_{i}(t)$ (see Figure~\ref{neura}). The activation state $A_{i}(t)$ of neurone $i$ at time $t$---a
bounded real or integer number---is then given by the neuron's \emph{
	activation function} $h_{i}$ such that $A_{i}(t)=h_{i}(U_{i}(t)).$
Typically, $h_{i}$ is either a linear function, a non-linear (step)
function, or a sigmoid function (e.g.: $tanh(x)$). In addition, $\theta _{i}$ (an extra weight with input always fixed at $1$) is known as the \emph{threshold} of neurone $i$. We say that neurone $i$ is \emph{active} at time $t$
if $A_{i}(t)>\theta _{i}.$ Finally, the neurone's output value $O_{i}(t)$ is
given by its output function $f_{i}(A_{i}(t))$. Usually, $f_{i}$ is the
identity function.

The units of an ANN can be organised in layers. A \emph{n-layer
	feedforward network} is an acyclic graph. It consists of a sequence of
layers and connections between successive layers, containing one input
layer, $n-2$ hidden layers, and one output layer, where $n\geq 2$. When $n=3$, we say that the network is a \emph{single hidden layer network}. When each
unit occurring in the $i$-$th$ layer is connected to each unit occurring in
the $i+1$-$st$ layer, we say that the network is \emph{fully-connected}.

A multilayer feedforward network computes a function $\varphi :
\mathbb{R}
^{r}\rightarrow 
\mathbb{R}
^{s}$, where $r$ and $s$ are the number of units occurring, respectively, in
the input and output layers of the network. In the case of single hidden
layer networks, the computation of $\varphi $ occurs as follows: at time $t_{1}$, the input vector is presented to the input layer. At time $t_{2}$,
the input vector is propagated through to the hidden layer, and the units in
the hidden layer update their input potential and activation state. At time $t_{3}$, the hidden layer activation state is propagated to the output layer,
and the units in the output layer update their input potential and
activation state. At time $t_{4}$, the output vector is read off the output
layer. In addition, most neural models have a \emph{learning rule},
responsible for changing the weights of the network progressively so that it
learns to approximate $\varphi $ given a number of \emph{training examples}
(input vectors and their respective target output vectors).

In the case of \emph{backpropagation}---probably the most commonly applied 
neural learning algorithm \citep{rumelhart}---an \emph{error} is
calculated as the difference between the network's actual output vector and
the target vector, for each input vector in the set of examples. This error $\mathbf{E}$ is then propagated back through the network, and used to
calculate the variation of the weights $\triangle \mathbf{W}$. This
calculation is such that the weights vary according to the \emph{gradient}
of the error, i.e. $\triangle \mathbf{W}=-\eta \nabla \mathbf{E},$ where $
0<\eta <1$ is called the \emph{learning rate}. The process is repeated a
number of times in an attempt to minimise the error, and thus approximate
the network's actual output to the target output, for each example. In order
to try and avoid shallow local minima in the error surface, a common
extension of the learning algorithm above takes into account, at any time $t$, not only the gradient of the error function, but also the variation of the
weights at time $t-1$, so that $\triangle \mathbf{W}_{t}=-\eta \nabla 
\mathbf{E}+\mu \triangle \mathbf{W}_{t-1}$, where $0<\mu <1$ is called the 
\emph{term of momentum}. Typically, a subset of the set of examples
available for training is left out of the learning process so that it can be
used for checking the network's \emph{generalisation} ability, i.e. its
ability to respond well to examples not seen during training.

CILP now is a neural-symbolic system based
on an ANN that integrates inductive learning and
deductive reasoning. In CILP, a \emph{translation algorithm} maps a logic
program $\mathcal{P}$ into a single hidden layer ANN $\mathcal{N}$
such that $\mathcal{N}$ computes the least fixed-point of $\mathcal{P}$ \citep
{lloyd}. This provides a massively parallel model for computing the stable
model semantics of $\mathcal{P}$ \citep{[GelfondLifschitz88]}. In addition, $\mathcal{N}$ can be trained with examples using a neural learning algorithm, having $\mathcal{P}$ as background knowledge. The
knowledge acquired by training can then be extracted \citep{GarBroGab01},
closing the learning cycle, as advocated by \cite{towell2}.

Let us exemplify how CILP\emph{'s} \emph{translation algorithm} works. Each
rule ($r_{l}$) of $\mathcal{P}$ is mapped from the input layer to the output
layer of $\mathcal{N}$ through one neurone ($N_{l}$) in the single hidden
layer of $\mathcal{N}$. Intuitively, the \emph{translation algorithm} from $\mathcal{P}$ to $\mathcal{N}$ has to implement the following conditions: (c$_{1}$) the input potential of a hidden neurone $N_{l}$ can only exceed its
threshold $\theta _{l}$, activating $N_{l}$, when all the positive
antecedents of $r_{l}$ are assigned truth-value $true$ while all the
negative antecedents of $r_{l}$ are assigned $false$; and (c$_{2}$) the
input potential of an output neurone $A$ can only exceed its threshold ($\theta _{A}$), activating $A$, when at least one hidden neurone $N_{l}$ that
is connected to $A$ is activated.

\begin{figure}[htb]
	\centerline{\includegraphics[height=5.5cm]{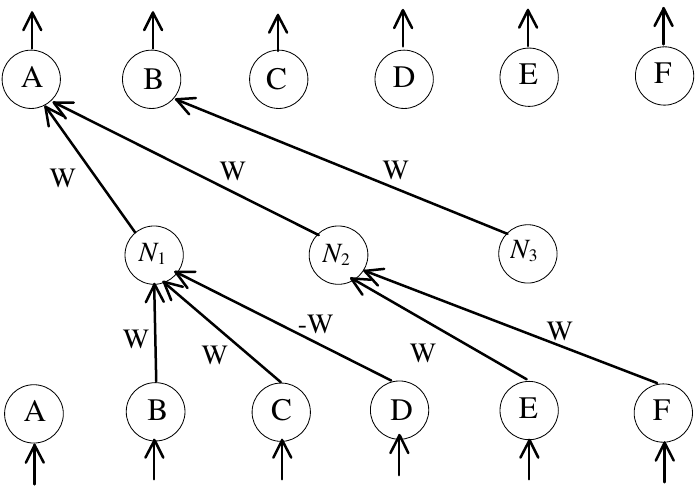}}
	\caption{A neural network for logic program $\mathcal{P}$.}
	\label{cilpex1}
\end{figure}

\begin{example}
	(CILP)\label{examp1} Consider the logic program $\mathcal{P}=\{B\wedge
	C\wedge \sim D\rightarrow A, E\wedge F\rightarrow A, B\}$, where $\sim$
	stands for LP's \emph{\ negation by failure} (a.k.a. default negation) \citep{lloyd}. Given $\mathcal{P}$, the
	CILP \emph{translation algorithm} produces the network $\mathcal{N}$ of
	Figure~\ref{cilpex1}, setting weights ($W$) and thresholds ($\theta $) in a
	way that conditions (c$_{1}$) and (c$_{2}$) above are satisfied. Note that,
	if $\mathcal{N}$ ought to be fully-connected, any other link (not shown in
	Figure~\ref{cilpex1}) should receive weight zero initially. Each input and
	output neurone of $\mathcal{N}$\ is associated with an atom of $\mathcal{P}$.
	As a result, each input and output vector of $\mathcal{N}$ can be associated
	with an interpretation for $\mathcal{P}$. Note also that each hidden neurone $
	N_{l}$ corresponds to a rule $r_{l}$ of $\mathcal{P}$ such that neurone $N_{1}
	$ will be activated if neurones $B$ and $C$ are activated while neurone\ $D$
	is not; output neurone $A$ will be activated if either $N_{1}$ or $N_{2}$ is
	activated; and output neurone $B$ will be activated if $N_{3}$ is, while $
	N_{3}$ is always activated regardless of the input vector (i.e. $B$ is a 
	\emph{fact}). To compute the stable models of $\mathcal{P}$, the output
	vector is recursively given as the next input to the network such that $
	\mathcal{N}$ is used as a recursive network to iterate the fixed-point operator of $\mathcal{P}$ 
	as suggested by \cite{garcez_book}. For example, output neurone $B$ should feed input neurone $
	B$. $\mathcal{N}$ will eventually converge to a stable state which is
	identical to the stable model of $\mathcal{P}$ provided that $\mathcal{P}$
	is an acceptable program \citep{AptPedreschi93}. For example, given any
	initial activation in the input layer of $\mathcal{N}_{r}$ (i.e. the network of
	Figure~\ref{cilpex1} recurrently connected), it always converges to a stable
	state in which neurone $B$ is activated and all the other neurones are not. We
	associate this with literal $B$ being assigned truth-value $true$, while all
	the other literals are assigned truth-value $false$, which represents the
	unique fixed-point of $\mathcal{P}$.
\end{example}

\emph{CILP} thereby provides a (provably sound) translation from a symbolic representation into an ANN that can be trained with examples as part of a knowledge evolution process, whereby the original symbolic representation is seen as background knowledge to the network. In what follows, we extend CILP to handle a range of normative rules and prove soundness. Notice how in the standard CILP translation, the weights of the connections linking the hidden and output layers of the network are always positive. As will become clearer in what follows, normative rules require the use of negative weights from the hidden to the output layer of the CILP network as well. This implements priorities in the rules \citep{garcez_book} and is responsible for adding alternative paths that enable robustness in the networks also. As we study such different forms of representation in different applications, such as CTD, we are interested in proving soundness, but also in efficient computation and learning, as exemplified later in this article.

\subsection{Input/Output Logic}\label{i/o}

As explained by \cite{makinson:whatisiol}, I/O logic takes its origin in the study of conditional norms which, either in imperative or indicative form, express obligations under some legal, moral, or practical code, goals, contingency plans, advice, etc. Putting this overall notion in formal terms, \cite{makinson2000input} represent rules  by ordered pairs $(a,x)$, where the antecedent~$a$ is thought of as an input, representing some condition or situation, and the consequent~$x$ is thought of as an output, representing what the rule tells us to be desirable, obligatory, or whatever else in that situation. 

Concerning the overall motivation behind the development of I/O logic, in philosophy---but also significant in our current context---norms are commonly distinguished from declarative statements. The latter may bear truth-values, while describing norms as true or false is meaningless. Instead, norms may be respected (or not), can be in force in the current context (or not), or can be assessed from the standpoint of other norms (e.g., when judging a law from a moral point of view). Still, much work addressing deontic formalisms in the study of logic and AI seem to ignore this distinction: most presentations of deontic logic---whether axiomatic or semantic---treat norms as if they could be subjected to an assessment in terms of truth-values. In particular, the truth-functional connectives `and', `or', and `not' are routinely applied to norms, forming compound norms out of elementary ones. Semantic constructions using possible worlds go further by offering rules to determine, in a model, the truth-value of a norm. I/O logic has its source in precisely this tension between philosophy and studies in formal logic (the reader may identify a similar tension between human reasoning and formal classical logic, as discussed earlier in the case of discourse processing). 

In the following, we first present abstract normative systems as a general descriptive framework for formal approaches to normative reasoning and basis for the subsequent introduction of propositional I/O logic, both of which then are applied to modelling the three types of permissions commonly encountered in normative context.

\subsubsection{Abstract Normative Systems}\label{ANS}

Modal logic has been the standard for normative reasoning ever since \cite{wright:51}. Still, for instance in \cite{handbook_deontic_logic}'s ``Handbook of Deontic Logic and Normative Systems'', the classical modal logic framework is mainly confined to the historical chapter. Another chapter presents the alternatives to the modal framework, and three chapters discuss concrete approaches, namely I/O logic, the imperativist approach \citep{H06}, and the algebraic conceptual implication structures \citep{lindahl:03}.\footnote{Of course this list is non-exhaustive as there are further alternative candidates for a new standard, such as nonmonotonic logic \citep{horty} or deontic update semantics \citep{torre:99}.} Against this multitude of approaches as backdrop, \cite{DBLP:conf/kr/TosattoBTV12} proposed abstract normative systems as common framework for comparing and analysing these new proposals.

Abstract normative systems study frameworks such as I/O logic on a general level, to which \cite{DBLP:conf/kr/TosattoBTV12} add two notions. First, each element in the (finite) universe comes with its ``anti-element'': this is the minimal extension to represent violations, namely elements in the input whose anti-element is in the output. Second, there is an element in the universe called~$\top$, contained in every context. 

\begin{definition}[Universe $L$ \citep{DBLP:conf/kr/TosattoBTV12}]
	Given a {\em finite} set of atomic elements $E$, the universe $L$ is $E\cup \{\simm e\mid e\in E\}\cup\{\top\}$. For $e\in E$, let $\overline{a}=\simm e$ iff $a=e$, $\overline{a}=e$ iff $a=\simm e$, and undefined iff $a=\top$.
\end{definition}
An abstract normative system is a directed graph, and a context is a set of nodes of the graph containing $\top$. 
In abstract normative systems there are three kinds of relations, for the regulative, permissive, and constitutive norms, respectively. We start with the regulative norms only.
The edges in an abstract normative system exactly define what a ``conditional norm'' (with respect to this abstract normative system) is.

\begin{definition}[ANS $\langle L,N\rangle$ \citep{DBLP:conf/kr/TosattoBTV12}]
	An abstract normative system ANS is a pair $\langle L,N\rangle$ with $N\subseteq L\times L$ a set of pairs of the universe, called conditional norms, and $A\subseteq L$ a subset of the universe such that $\top\in A$, called the context.
\end{definition}

In a context, an abstract normative system generates or produces an obligation set, a subset of the universe, reflecting the obligatory elements of the universe. 
The class of deontic operations is specified by their domain and codomain. Some examples of deontic operations are given below.

\begin{definition}[Deontic operation $\out$ \citep{DBLP:conf/kr/TosattoBTV12}]
	A deontic operation $\out$ is a function from an abstract normative system $\langle L,N\rangle$ and a context $A$ to a subset of the universe $\out(\langle L,N\rangle,A)\subseteq L$.
	Since $L$ is always clear from context, we write $\out(N,A)$ for $\out(\langle L,N\rangle,A)$.
\end{definition}

Simple-minded output or $\out_1$ is Makinson and van der Torre's minimal system. 
Basic output or $\out_2$ allows for reasoning by cases, which now means that if something is obligatory in the context of $a$ and its complement $\overline{a}$, then it is obligatory also in the minimal context.
Reusable output or~$\out_3$ allows for deontic detachment, which now corresponds to iteration of the rules.
Throughput or $\out_i^+$ allows for identity. 
All possible combinations lead to eight input/output operations.

\begin{definition}[Eight deontic operations \citep{DBLP:conf/kr/TosattoBTV12}]\label{def:unconstrained}
	A context $A\subseteq L$ is complete if for all $e\in E$, it contains either $e$ or $\overline{e}$ (or both). 
	\begin{description}
		\item $\out_1 (N,A) = N(A)=\{x \mid (a,x)\in N \mbox{ for some }a\in A\}$
		\item $\out_2 (N,A) = \cap \{ N(V) \mid A \subseteq V, V$ complete$\}$
		\item $\out_3 (N,A) = \cap \{ N(B) \mid A \subseteq B \supseteq N(B) \}$
		\item $\out_4 (N,A) = \cap \{N(V) \mid A \subseteq V \supseteq N(V), V \mbox{ complete}\}$
		\item $\out_i^+ (N,A) = \out_i(N\cup \{(a,a)\mid a\in L\},A)$
	\end{description}
	Equivalently, $\out_3(N,A)$ can be defined as $N(B)$ where $B$ is the {\em smallest} set containing $A$ and closed under $N$, i.e. \mbox{$A \subseteq B \supseteq N(B)$}. Moreover, to emphasise symmetry, $\out_1 (N,A)$ can be defined equivalently as $ \cap \{ N(B) | A \subseteq B\}$.
\end{definition}

At least since the work of \cite{horty}, nonmonotonic techniques have been used to deal with reasoning in the context of dilemmas, CTD reasoning, and defeasible norms:
\begin{itemize}
	\item \emph{Dilemmas} are two (or more) obligations with contradictory content, like the obligation for $a$ and the obligation for $\overline{a}$.
	\item \emph{CTD} or secondary obligations $(a,x)$ are in force only in case of {\em violation} of a primary obligation, e.g., generated using $(\top,\overline{a})$.\footnote{To give an intuitive example of a CTD, we report the so-called {\em dog-sign} example by \cite{Prakken97} already hinted at in the introduction: ``Suppose that: there must be no dog around the house, and if there is no dog, there must be no warning sign, but if there is a dog, there must be a warning sign.'' Obviously, if there is a dog, the conditional obligation that there must be no sign does not become unconditional, since its condition is not fulfilled. On the other hand, it can also be inferred that if no obligations are violated, there will be no sign (modulo exceptions, of course).}
	\item \emph{Defeasible deontic logic} is concerned with violations {\em and exceptions}  \citep{torre:thesis,nute:ddl}.
\end{itemize}

\subsubsection{Propositional Input/Output Logic}

As explained by \cite{makinson2000input}, propositional I/O logic establishes a relatively simple setting, abstracting from important aspects of deontic reasoning, such as CTD reasoning or permissions.\footnote{These require much more involved I/O operations, which we shortly discuss in Section~\ref{normativepriorities} below. Cf. the work by \cite{makinson:ciol} and \cite{makinson:perm} for more detailed treatments.} The construction of the semantics is analogous to the just discussed abstract normative systems, adding the closure of input and output under propositional consequence.
As before, $N(A)=\{x \mid (a,x)\in N \mbox{ for some }a\in A\}$.

\begin{definition}[\textit{out} \citep{makinson2000input}] \label{obligation:definition}
	Let $L$ be a propositional logic with $Cn$ the consequence 
	operator of $L$, $\top$ a tautology of $L$, a \textit{complete set} one that is 
	either maxiconsistent or equal to $L$, and let $N$
	be a set of ordered pairs of $L$ (called the generators). A generator
	$(a,x)$ is read as `if input $a$ then output $x$'. The following
	logical systems are defined:\\
	$out_1(N,A) =  Cn(N(Cn(A))$
	\\
	$out_2(N,A) = \cap\{Cn(N(V)):A\subseteq V,V \mbox{ complete}\}$
	\\
	\mbox{$out_3(N,A) = \cap\{Cn(N(B)):A\subseteq B= Cn(B)\supseteq N(B)\}$}
	\\
	\mbox{$out_4(N,A) = \cap\{Cn(N(V)):A\subseteq V\supseteq N(V), V \mbox{ complete}\} $}
\end{definition}

Note that neither in the I/O logic framework, nor in the abstract normative systems framework, does a normative system `imply' a norm. Norms are used to generate obligation sets; we can axiomatise deontic operations using a proof system based on conditionals, but this does not mean that norms are ``implied'' or ``derived.'' The most we can say is that a norm is ``accepted'' by a normative system \citep{torre:99}, or ``redundant'' in a normative system \citep{torre:deon10}.
The latter point may be related to two philosophical considerations of the I/O logic framework. First, as already explained above, the framework is based on the idea that norms do not have truth values, known as J\"{o}rgensen's dilemma in the deontic logic literature \citep{jorgensen}. Second, 
the role of logic is not to create or determine a distinguished set of norms, but rather to prepare
information before it goes in as input to such a normative code, to unpack output as it emerges and, if needed, coordinate the two in certain ways.  
A set of conditional norms is thus seen as a transformation device, and the task of logic is to act as its ``secretarial assistant'' \citep{makinson2000input}. 

\subsection{Overview of the Architecture}\label{architecture}

Our goal is to allow the agent to learn about norms and their interpretation from experience, and to take decisions which respect the norms she is subject to at the respective point in time. Thus, the agent needs to know what is obligatory and forbidden according to norms (conditional rules) in any situation in real time: what is obligatory can eventually become an action of the agent, while what is forbidden inhibits such actions. 
Also, rules may change as the normative environment changes over time. The agent should be flexible enough to adapt her behaviour to the context using as information the instances of behaviours which have been considered illegal.

\begin{figure}[h]
	\centering
	\hspace*{-4mm}
	\includegraphics[scale=0.45]{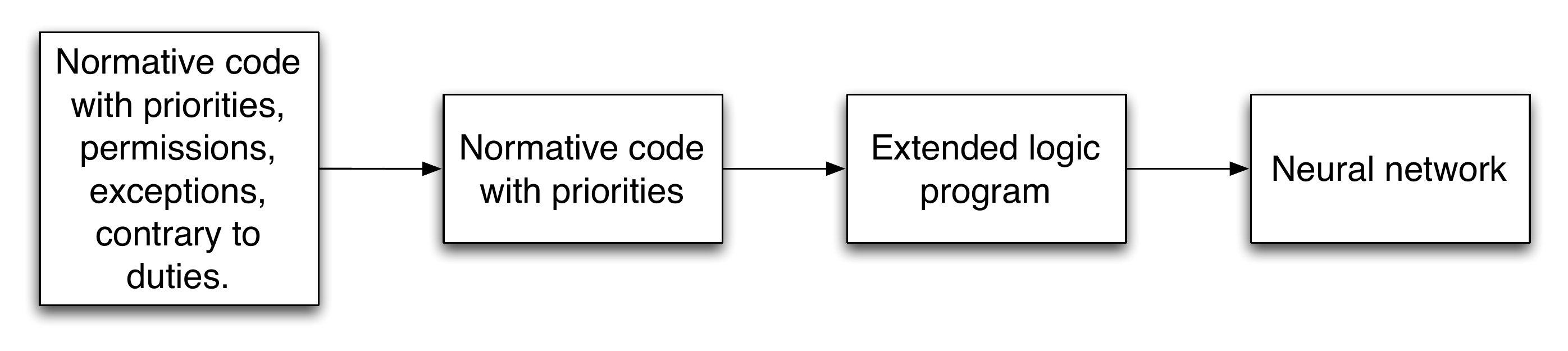}
	\caption{From normative codes to neural networks.}\label{fig:flow}
\end{figure}

To allow an intelligent agent to have a internal representation of a normative code, we follow the process visualised in Figure~\ref{fig:flow}. The encoding process is a single and unique task, and we just decompose it in subtasks to give a more detailed explanation. The first step involves encoding a list of normative aspects in terms of priorities and will be described in the next subsection. The second step translates a normative code in I/O logic into an extended logic program (i.e. LP extended with \emph{classical negation}, a.k.a. explicit negation, which leads to the answer set semantics of LP mentioned earlier in the context of three-valued logics. The third step applies a translation algorithm to convert the logic program into a neural network. The last two steps will be analysed in detail in Section~\ref{ncil}. \\

\begin{figure}[h]
	\centering
	\includegraphics[scale=0.6]{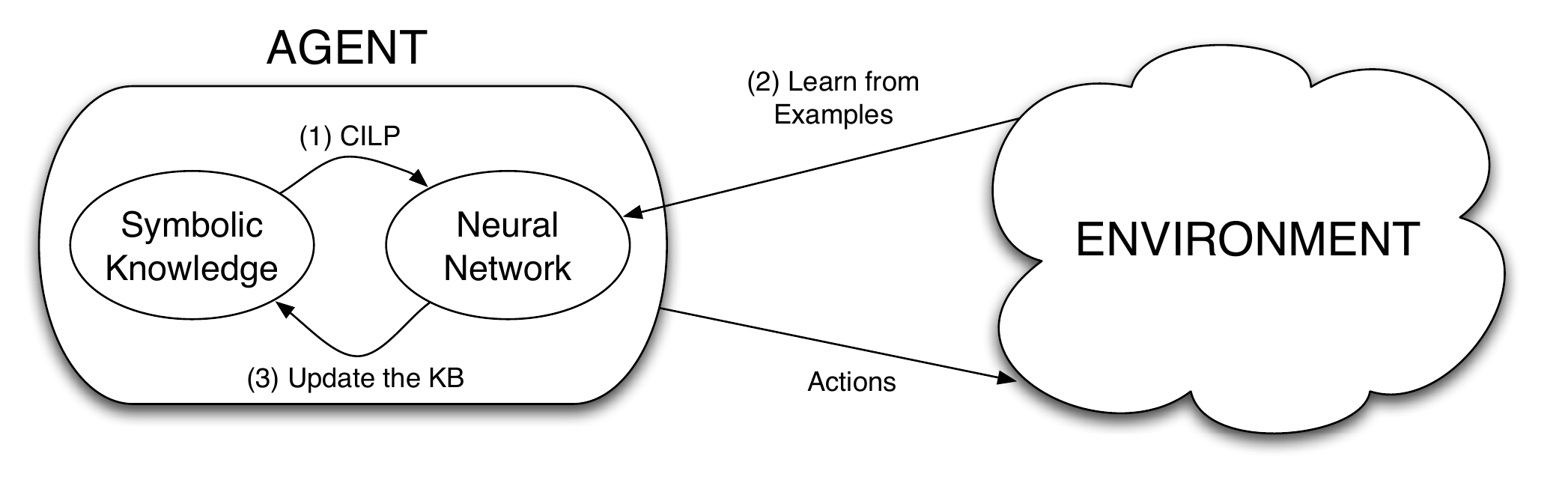}
	\caption{Normative agent architecture.}\label{fig:normAgent}
\end{figure}

Figure~\ref{fig:normAgent} describes our approach from a more abstract perspective. Note that the encoding of a normative code in an ANN is lumped to a single step. Our framework starts from the symbolic KB of norms contained in the agent, transforming it into an ANN using the encoding introduced in Figure~\ref{fig:flow} and described below. The ANN is structured as follows: input neurones of the network represent the state of the world, while the output neurones represent the obligations of the agent, or the prohibitions. The ANN is used as part of the controller for the agent and, given its ability to change (i.e. learn from examples), it is expected to give the agent the required flexibility.

\subsection{Normative Problems as Priorities}\label{normativepriorities}

Recall, as discussed, that normative reasoning requires agents to deal with specific problems such as \emph{dilemmas}, \emph{exceptions}, and \emph{CTDs}. 
In what follows, a norm N will be expressed as labelled generators $N=(I,O)$, read `if input $I$ then output $O$'. In general, $I$ in $(I,O)$ is any propositional formulae, which will be restricted later to conjunctions of literals. \\ 

{\bf Dilemmas:} two obligations are said to be \emph{contradictory} when they cannot be accomplished together. A possible example of contradictory norms is the \emph{dilemma}. This usually happens when an agent is subject to different normative codes (i.e. when an agent has to follow the \emph{moral} and the \emph{legal} code). How to overcome dilemmas is left as future work, as we are focusing on how to use \emph{priorities} to regulate \emph{exceptions} and \emph{CTDs}.\\

{\bf Priorities} are used to give a partial ordering between norms. This is useful when, given two applicable norms, we always want one to preempt the other, for instance when dealing with \emph{exceptions}.
We encode priorities among the norms by using \emph{negation by failure} ($\sim$). Given two norms $N_1 = (A_1 \wedge A_3, \beta_1)$ and $N_2 = (A_2 \wedge A_3, \beta_2)$ and a priority relation $N_1 \succ N_2$ between the norms (such that the first norm has priority), we encode the priority relation by modifying the antecedent of the norm with lower priority. Specifically, we include in the antecedent of the norm with the lower priority the negation-as-failure of the literals in the antecedent of the higher priority norm that does not appear in the antecedent of the lower priority norm. We do so in order to ensure that, in a situation where both (unmodified) norms would be applicable, the newly inserted negation-as-failure atoms in the antecedent of the modified lower-priority norm evaluate to $false$ and make the norm not applicable. Considering for example the two norms given above, we have to modify $N_2$. The only atom appearing in $N_1$'s input and not in $N_2$'s input is $A_1$, and therefore we introduce $\sim A_1$ as a conjunct in $N_2$'s input. After embedding the priority, the second norm becomes $N_2' = (A_2 \land \sim A_1 \land A_3, \beta_2)$. Note that in a potentially conflicting situation when $A_1$, $A_2$ and $A_3$ hold, $N_1$ and $N_2$ are applicable, but $N_2'$ is not, thus avoiding the conflict.\\

{\bf Exceptions} occur when, due to particular circumstances, a norm should be followed instead of another. Suppose that a norm $N_3 = (\alpha, \beta)$ should be applied in all the situations containing $\alpha$. For exceptional situations we consider an additional norm $N_4 = (\alpha \land \gamma, \neg \beta)$. The latter norm should be applied in a subset of situations w.r.t. $N_3$: specifically all those when, in addition to $\alpha$, also $\gamma$ holds. We can call situations where both $\alpha$ and $\gamma$ hold exceptional situations. In these exceptional situations both norms could be applied. This would produce two contrasting obligations: $\beta$ and $\neg \beta$. To avoid this we add the following priority relation: $N_4 \succ N_3$. Therefore we modify the input of the norm with lower priority as described earlier. The result is a new norm $N_3' = (\alpha \land \sim \gamma, \beta)$, that would not be applied in the exceptional situations, avoiding the problem of contrasting obligations.\\ 

{\bf CTDs:} An important property of norms is that they are soft constraints and, accordingly, can be violated. CTDs provide additional obligations to be fulfilled when a violation occurs. For example, consider a norm $N_5 = (\alpha, \beta)$ that should be applied in all situations containing $\alpha$ and producing the obligation $\beta$. As mentioned, norms can be violated, therefore we can also define a norm that produces alternative obligations to be followed in case of a violation. Let this new norm be $N_6 = (\alpha \land \neg \beta, \gamma)$. The latter norm contains in its input both the input of $N_5$ and the negation of its output. In this way it describes which should be the alternative obligation to $\beta$ in the case that it cannot be achieved, in this example $\gamma$.
We use a priority relation between the two norms in order to avoid the generation of the obligation $\beta$ in case it is already known that it is not satisfiable. We add then the following priority relation $N_6 \succ N_5$ that modifies the first norm as follows:  $N_5' = (\alpha \land \sim \neg \beta, \beta)$.\\ 

{\bf Permissions:} An important distinction between \emph{obligations} and \emph{permissions} is that the latter will not be explicitly encoded in the ANN. In our approach we consider that something is permitted to the agent if not explicitly forbidden (note that we consider the ought of a negative literal as a prohibition). Due to this, we assume that norms with a permission in their output implicitly have priority over the norms that forbid the same course of action\footnote{\cite{makinson:perm} consider three kinds of permissive norms, namely negative, positive, and static positive permission. In this article, we restrict discussions to the above, and should note that much future work is left to be done when it comes to the provision of connectionist representations for normative and deontic reasoning systems}. For example, using $P$ in the output of a norm to denote a \emph{permission}, consider two norms $N_7 = (A_1, P(\beta_1))$, $N_8 = (A_2, \neg \beta_1)$. The first norm permits $\beta_1$ and the second forbids it. In this case, we use the following priority relation: $N_7 \succ N_8$. 

\subsection{Normative Connectionist Inductive Learning and Logic Programming}\label{ncil}

In this section we introduce a new approach for coding a fragment of I/O logic which corresponds to extended LP into ANNs. The main intuition is that, although logic programs in general do not explicitly capture the concepts of \emph{inputs} and \emph{outputs}, a neural-symbolic system based on extended logic programming does - on a purely structural level: inputs and outputs in I/O logic correspond to the input and output layers of the ANN - and allows the representation of norms in ANNs.   

As described above, in I/O logic norms are represented as ordered pairs of formulas like $(\alpha,\beta)$. A peculiarity of I/O logic is that it does not have $(\alpha,\alpha)$ for any $\alpha$ (i.e. \emph{identity} is not an axiom). 
In normative reasoning, the input does not necessarily become an output: the reason is that the output is interpreted as what is obligatory, thus, just because $a$ is in the input, it is not necessarily the case that $a$ is obligatory as well. This I/O perspective corresponds straightforwardly to the general intuition behind an ANN. Activating input neurone $A$ in Figure~\ref{cilpex1} does not necessarily activate output neurone $A$ also; this is true for any neurone, and it allows a subtle but important distinction between the activation of an input neurone which is derived from the context, that is, the input values provided to the network, and the activation of an output neurone, which is derived from the KB. For example, in Figure~\ref{cilpex1}, the truth-value of $B$ in the input is, at first, obtained from the input to the network (its context), whilst the truth-value of $B$ in the output is $true$ ($B$ is a fact in the KB).
Modifying the original CILP algorithm, we first translate  I/O logic into an extended logic program to be processed by CILP without requiring inputs to be always translated  into outputs as well, so that the ANN is allowed different input and output layers. The input $\alpha$ of an I/O norm $(\alpha,\beta)$ is subsequently passed as an input vector to the network, producing an output representing what is obligatory (e.g. $\beta$, if the translation to the ANN is proved correct). Only some input appears in the output, if it is made obligatory by a norm.
In CILP, output nodes are always connected to input nodes creating a recurrent network, to represent the transitivity of logical rules when computing minimal or stable models. In normative reasoning, transitivity is not always accepted (since if you are obliged to do $a$ and, if $a$ then you are obliged to do $b$ does not imply that you are obliged to do $b$). Thus, the normative CILP extends CILP also to allow that certain outputs might not be connected to their corresponding inputs (or will not even have a corresponding input as a result of the first change made to CILP earlier).

\subsubsection{Mapping Input/Output Logic into Neural Networks}
\label{mappingIO}
We now first introduce a specific fragment of I/O logic relevant for our purposes, then we present an embedding of this fragment into extended logic programs, and finally,  how to represent such norms with priorities in ANNs.

\begin{definition}\label{elp}
	An \emph{extended logic program} is a finite set of clauses of the form $L_0 \leftarrow L_1, \ldots,\sim L_n, \sim L_{n+1}, \ldots, \sim L_{m}$, where $L_i$ ($0 \leq i \leq n$) is a literal i.e. an atom or 
	a \emph{classical negation} of an atom denoted by $\neg$, and $\sim L_j$ ($n+1 \leq j \leq m$) is called \emph{default literal}, where $\sim$ represents negation-as-failure. Following \cite{[GelfondLifschitz88]}, from now on we use `$\leftarrow$' in place of `$\rightarrow$', and say that $L_0$ is $true$ if $L_1,...,L_m$ is $true$ ($L_0 \leftarrow L_1,...,L_m$), where $L_1,...,L_m$ denotes a conjunction of literals (with `,' used in place of `$\wedge$'). 
\end{definition}
Given an extended logic program $P$ we identify its \emph{answer sets} \citep{gelfond} as $EXT(P)$.
\begin{definition}[I/O Normative Code]\label{iorules}
	A normative code $\mathbf{G} = \langle \mathbb{O}, \mathbb{P}, \succ \rangle$ is composed by two sets of rules $r:(\alpha,\beta)$ and a preference relation 
	$\succ$ among those rules. 
	Rules in $\mathbb{O}$
	are called \emph{obligations}, while rules in $\mathbb{P}$ are \emph{permissions}. Rules in $\mathbb{O}$ are of the type $(\alpha,\beta)$, where:
	\begin{itemize}
		\item $\alpha = \alpha_1 \vee \ldots \vee \alpha_n$ is a propositional formula in disjunctive normal form, i.e. $\alpha_i$ (for $0 \leq i \leq n$) is a conjunction of literals
		$(\neg a_{\alpha_{i1}} \wedge \ldots \wedge \neg a_{\alpha_{im}} \wedge a_{\alpha_{i(m+1)}} \wedge \ldots \wedge a_{\alpha_{1(m+p)}})$. Without loss of generality
		we assume that the first $m$ literals are negative while the other $p$ are positive.
		\item $\beta = \neg b_{\beta_{1}} \wedge \ldots \wedge \neg b_{\beta_{m}} \wedge b_{\beta_{m+1}} \wedge \ldots \wedge b_{\beta_{m+p}}$ is a finite conjunction of literals.
	\end{itemize}
	Rules in $\mathbb{P}$ are of type $(\alpha,l)$, where $\alpha$ is the same as for obligations, but $l$ is a literal.
\end{definition}
As put forward by \cite{BoellaT05}, one of the roles of permissions is to undercut obligations. Informally,
suppose to have a normative code $\mathbf{G}$ composed of two rules:
\begin{enumerate}
	\item $b$ is obligatory (i.e. $(\top,b) \in \mathbb{O}$).
	\item If $a$ holds, then $\neg b$ is permitted (i.e. $(a,\neg b) \in \mathbb{P}$).
\end{enumerate}
We say that  the rule $(a, \neg b)$ \emph{has priority} over $(\top,b)$, i.e. $b$ is 
obligatory as long as $a$ does not hold, otherwise $\neg b$ is permitted and, therefore
$b$ is not obligatory anymore. 

The fact that we consider only the I/O rules as introduced in Definition~\ref{iorules} permits us to give a natural embedding of this fragment of I/O logic into extended logic programs.

\begin{definition}\label{def_fun_trans}
	Let $\lceil \cdot \rceil$ denote a function mapping I/O rules (Definition~\ref{iorules}) into extended logic programs (Definition~\ref{elp}), as follows:
	\[
	\begin{array}{c}
	\lceil r: (\alpha_1 \vee \ldots \vee \alpha_n, \beta_1 \wedge \ldots \wedge \beta_m) \rceil = \\
	\{ r_{11}:(\lceil \beta_1 \rceil_{out} \leftarrow \lceil \alpha_1 \rceil_{in}); \ldots; r_{1m}:(\lceil \beta_m \rceil_{out} \leftarrow \lceil \alpha_1 \rceil_{in}); \ldots;\\
	r_{n1}:(\lceil \beta_1 \rceil_{out}  \leftarrow \lceil \alpha_n\rceil_{in}); \ldots; 
	r_{nm}:(\lceil \beta_m \rceil_{out} \leftarrow \lceil \alpha_n \rceil_{in})\}\\
	\\
	\lceil l_1 \wedge \ldots \wedge l_n \rceil_{in/out} = \lceil l_1 \rceil_{in/out}, \ldots,  \lceil l_n \rceil_{in/out}\\
	\lceil a \rceil_{in} = in\_a \qquad \qquad \lceil a \rceil_{out} = out\_a \\
	\lceil \neg a \rceil_{in} = \neg in\_a \qquad \qquad \lceil \neg a \rceil_{out} = \neg out\_a
	\end{array}
	\]
	We call rules $r_{ij}$ instances of $r$, and we informally write $r_{ij} \in Ints(r)$.
\end{definition}
Notice that the program resulting from the application of $\lceil \cdot \rceil$ has a unique model because it is negation-as-failure free.
\begin{lemma} 
	Given a set of obligations $\mathbb{O} = \{(\alpha_1,\beta_1), \ldots, (\alpha_n,\beta_n)\}$. Then it holds that:
	\[
	\begin{array}{l}
	\mbox{ If } (\alpha,\beta) \in \mathbb{O} \mbox{ then } \lceil \beta \rceil_{out} \in 
	EXT(\{ \lceil (\alpha_1,\beta_1) \rceil; \ldots; \lceil (\alpha_n,\beta_n) \rceil \} \cup \lceil \alpha \rceil_{in}).
	\end{array}
	\]
	\begin{proof}
		The \emph{if }direction is trivial while the \emph{only if} can be proven by showing that every application of the immediate consequence operator $\mathcal{T}$ (as defined by \cite{gelfond}) can be encoded into an application of $\lceil \cdot \rceil$ (Definition~\ref{def_fun_trans}).
	\end{proof}
\end{lemma}
We now show how to extend the preference relation $\succ$ w.r.t. rules generated with $\lceil \cdot \rceil$.
\begin{definition}
	Given a normative code $\mathbf{G} = \langle \mathbb{O},\mathbb{P}, \succ \rangle$ we define a transformation $Tr_o( \cdot)$ 
	such that $Tr_o(\mathbf{G}) = \langle \lceil \mathbb{O} \rceil, \mathbb{P}, \succ' \rangle$, where $\succ'$ is defined
	as follows: 
	$t_{ij} \succ' t'_{i'j'}$, for all $t_{ij} \in Inst(t)$ and $t'_{i'j'} \in Inst(t')$  for $t,t' \in \mathbb{O}$ such that $t \succ t'$.
\end{definition}
For this reason, for a given normative code $Tr_o(\mathbf{G})$, we introduce a further transformation $Tr_p( \cdot )$ as follows:
\begin{definition}
	Given a normative code $\mathbf{G_o} = Tr_o(\mathbf{G}) = \langle \lceil \mathbb{O} \rceil, \mathbb{P}, \succ' \rangle$ we define $Tr_p(\mathbf{G_o}) = \langle 
	\lceil \mathbb{O} \rceil, \mathbb{P}, \succ''  \rangle$, where $\succ''$ is defined as follows:
	For all $p: (\alpha,l) \in \mathbb{P}$, 
	$p \succ'' t_{ij}$, for all $t_{ij}:(\alpha,\neg l) \in \lceil \mathbb{O} \rceil$ 
\end{definition}

We now recall how to encode (metalevel) preference relations, which define a priority between LP rules into (object-level) extended logic programs \citep{nute}.  

\begin{definition}
	\label{priorities}
	(\textbf{Object-level Priorities}) Given a preference relation between $r_{i}$ and $r$ such that $r_{i}\succ r$ for $1\leq i\leq j$, replace the clause $r:L_{q+1} \leftarrow (L_{1},...,L_{p})$ with the clause $L_{q+1} \leftarrow (L_{1},...,L_{p},\sim L^{1}_{p+1},...,\sim L^{1}_{q},...,\sim L^{j}_{p+1},...,\sim L^{j}_{q})$,
	where $r_{i(1 \leq i \leq j )}:(L^{i}_{q+1} \leftarrow L^{i}_{p+1},...,L^{i}_{q})$.
\end{definition}

\begin{example}\label{normative_code}
	Take the following normative code: $$\mathbf{G}= {\langle \{r:(a,\neg b \wedge c)\},\{p:(d,b)\},\{ \} \rangle}.$$ Then
	$Tr_o(\mathbf{G}) = \{\langle r_{11}: (a, \neg b); r_{12}: (a,c) \},\{p:(d,b)\},\{\} \rangle$,
	and $Tr_p(Tr_o(\mathbf{G})) = \{\langle r_{11}: (a, \neg b); r_{12}:(a,c) \},\{p: (d,b) \}, \{p \succ r_{11} \rangle \}$.
\end{example}

For rules with permissions in the output, which are
of the form $p_i: L_{i_{m+1}} \leftarrow (L_{i_1}, \ldots, L_{i_n}, L_{i_{n+1}}, \ldots, L_{i_{m}})$, such that, for any other rule, $r: \neg L_{i_{m+1}} \leftarrow (L_{i_1}, \ldots, L_{i_n})$ 
(resulting from the application of $\lceil \mathbf{G} \rceil$), we impose $p_i \succ r$. 
As discussed, the role of permissions is to undercut obligations in $\lceil \mathbf{G} \rceil$, and permissions will not be encoded explicitly into the ANN (every output of the ANN counts as an obligation; something is permitted if the contrary is not obligatory, see Section~\ref{io2nn}).

\begin{lemma}\label{extension_identity}
	Let $P_{\succ }=\{r_{1},r_{2},...,r_{n}\}$ be an extended logic program with an
	explicit preference relation $\succ$. Let $P$ denote the translation of 
	$P_{\succ }$ into a program without $\succ $ (Definition \ref{priorities}). It follows that $EXT(P_{\succ })=EXT(P)$.
\end{lemma}

We are particularly interested in the translation of $P_{\succ}$ into $P$ because it is well-known 
that CILP networks will always compute the unique answer set of $P$, by converging to a unique stable state, provided that $P$ is well-behaved (i.e. locally stratified, or acyclic, or acceptable, cf. \cite{garcez_book}). This will be explored further in the next subsection. Before proceeding, let us use an example to illustrate what has been achieved so far. \\

\begin{example}(Translation of normative code into extended logic program)
	Consider the following normative code:
	\noindent
	\begin{center}
		\begin{minipage}[t]{0.5\textwidth}
			$r1: (a \vee b,{\bf O}(c))$\\
			$r2: (d \wedge e,{\bf O}(f))$\\
			$r3: (g,{\bf P}(\neg f))$ \\
			$r1 \succ r2$ 
		\end{minipage}
		\linebreak\\
	\end{center}
	\noindent
	First, obligations are decomposed into instances: \\
	\noindent
	\begin{minipage}[t]{0.5\textwidth}
		{$r1: (a \vee b,{\bf O}(c))$}\\
		$r2: (d \wedge e,{\bf O}(f))$\\
		$r3: (g,{\bf P}(\neg f))$ \\
		$r1 \succ r2$
	\end{minipage}
	\begin{minipage}[t]{0.5\textwidth}
		\noindent
		$r1_1: c \leftarrow a$\\
		$r1_2: c \leftarrow b$\\
		$r2: f \leftarrow d,e$\\
		$r3: (g,{\bf P}(\neg f))$ \\
		$r1 \succ r2$
	\end{minipage}
	\linebreak\\
	
	\noindent
	Secondly, the priorities are decomposed: \\
	\noindent
	\begin{minipage}[t]{0.5\textwidth}
		$r1_1: c \leftarrow a$\\
		$r1_2: c \leftarrow b$\\
		$r2: f \leftarrow d,e$\\
		$r3: (g,{\bf P}(\neg f))$ \\
		{$r1 \succ r2$}
	\end{minipage}
	\begin{minipage}[t]{0.5\textwidth}
		\noindent
		$r1_1: c \leftarrow a$\\
		$r1_2: c \leftarrow b$\\
		$r2: f \leftarrow d,e$\\
		$r3: (g,{\bf P}(\neg f))$ \\
		{ $r1_1\succ r2$ \\
			$r1_2\succ r2$ \\ }
	\end{minipage}
	\linebreak\\
	
	\noindent
	Finally, the permission-generated priorities are added:\\
	\begin{minipage}[t]{0.5\textwidth}
		\noindent
		$r1_1: c \leftarrow a$\\
		$r1_2: c \leftarrow b$\\
		$r2: f \leftarrow d,e$\\
		$r3: (g,{\bf P}(\neg f))$ \\
		$r1_1\succ r2$ \\
		$r1_2\succ r2$ \\ 
	\end{minipage}
	\begin{minipage}[t]{0.5\textwidth}
		\noindent
		$r1_1: c \leftarrow a$\\
		$r1_2: c \leftarrow b$\\
		$r2: f \leftarrow d,e$\\
		$r3: (g,{\bf P}(\neg f))$ \\
		$r1_1\succ r2$ \\
		$r1_2\succ r2$ \\ 
		{ $r3\succ r2$ }\\
	\end{minipage}
	\linebreak\\
	
	\noindent
	And the priorities are encoded as norm inputs:\\
	\begin{minipage}[t]{0.5\textwidth}
		\noindent
		$r1_1: c \leftarrow a$\\
		$r1_2: c \leftarrow b$\\
		{ $r2: f \leftarrow d,e$\\ }
		$r3: (g,{\bf P}(\neg f))$ \\
		$r1_1\succ r2$ \\
		$r1_2\succ r2$ \\ 
		$r3\succ r2$
	\end{minipage}
	\begin{minipage}[t]{0.5\textwidth}
		\noindent
		$r1_1: c \leftarrow a$\\
		$r1_2: c \leftarrow b$\\
		{$r2: f \leftarrow d, e, \sim $\hspace*{-0.5mm}$a, \sim $\hspace*{-0.5mm}$ b, \sim $\hspace*{-0.5mm}$ g$\\}
	\end{minipage}
	\linebreak\\
	
	\noindent
	The result is an equivalent extended logic program.
\end{example}

\subsubsection{The N-CILP algorithm}\label{io2nn}
In this section we introduce the translation algorithm encoding a normative code into a feedforward ANN (with
semi-linear neurones), namely the Normative-CILP (N-CILP) algorithm. 
The proposed algorithm differs from CILP \citep{garcez_book} in how priorities are encoded into the ANN, and it does not assume identity.\\

\hrule\smallskip
\noindent\textbf{N-CILP Algorith} (Input: normative code $\mathbf{G}$; Output: ANN)
\begin{enumerate}
	\item $\mathbf{G'} = Tr_o(\mathbf{G}); G''=Tr_p(\mathbf{G'})$
	\item Apply the encoding of priorities as described in Definition~\ref{priorities} to $\mathbf{G''}$.
	\item For each rule $R_k = \beta_{o_1} \leftarrow \alpha_{i_1}, \ldots, \alpha_{i_n}, \sim \alpha_{i_n+1}, \ldots, \sim \alpha_{i_m} \notin \mathbb{P}$.
	\begin{enumerate}
		\item For each literal $\alpha_{i_j}$ ($1 \leq j \leq m$) in the input of the rule: if there is no input neurone labeled $\alpha_{i_j}$ in the input level,
		then add a neurone labeled $\alpha_{i_j}$ in the input layer.
		\item Add a neurone labeled $N_{k}$ in the hidden layer.
		\item If there is no neurone labeled $\beta_{o_1}$ in the output level, then add a neurone labeled $\beta_{o_1}$ in the output layer.
		\item For each literal $\alpha_{i_j}$ ($1 \leq j \leq n$): connect the respective input neurone with the neurone labeled $N_k$ in the hidden
		layer with a positive weighted arc.
		\item For each literal $\sim \alpha_{i_h}$ ($n+1 \leq j \leq m$): connect the respective input neurone with the neurone labeled $N_k$ in the hidden layer with a negative weighted arc (the connections between these input neurones and the hidden neurone of the rule represent the priorities translated with \emph{negation-as-failure}).
		\item Connect the neurone labeled $N_i$ with the neurone in the output level labeled $\beta_{o_1}$ with a positive weighted arc (each output in the rules is considered as a positive atom during the translation; a rule with a negative output $\lnot \beta$ is translated in the network as output neurone labeled $\beta '$ that has the same meaning of $\lnot \beta$ but for the purpose of the translation can be treated as a positive output).
	\end{enumerate}
\end{enumerate}
\hrule

\begin{proposition}
	For any normative code in
	the form of an extended logic program there exists an ANN
	obtained
	from the N-CILP translation algorithm such that the network computes the
	answer set semantics of the code.
	\begin{proof} 
		Definition~\ref{def_fun_trans} translates a normative code into an extended logic
		program having a single extension (or answer set). From Lemma~\ref{extension_identity}, the
		program extended with a priority relation also has a single extension.
		\cite{garcez_book} show that any extended logic program
		can be encoded into an ANN. N-CILP performs one such encoding
		using network weights as defined by \cite{garcez_book}. Hence,
		N-CILP is sound. Since the program has a single extension, the
		iterative recursive application of input-output patterns to the network
		will converge to this extension, which is identical to the unique
		answer set of the program, for any initial input.
	\end{proof}
\end{proposition}

We end this subsection with a complete example of a translation of a normative code to an ANN. The following captures parts of the rule set a soccer-playing agent might be equipped with regarding the need to stop an opponent from scoring a goal in different situations (as, for instance, potentially encountered in the RoboCup robot soccer competitions):\\
\hrule\smallskip
\small
\noindent R1 = (opponentShooting $\wedge$ closeToOpponent, {\bf O}(impactingOpponent)) \\
\noindent R2 = (goalkeeper $\wedge$ insideOwnArea $\wedge$ closeToOpponent $\wedge$ opponentHasBall,\\\noindent {\bf O}(impactingOpponent))\\
\noindent R3 = (haveBall $\wedge$ closeToGoal $\wedge$ closeToOpponent, {\bf O}(impactingOpponent))\\
\hrule\smallskip
\normalsize
This set of norms is translated to an extended logic program: \\
\hrule\smallskip
\small
\noindent impactingOpponent $\leftarrow$ opponentShooting $\wedge$ closeToOpponent  \\
\noindent impactingOpponent $\leftarrow$ goalkeeper, insideOwnArea, closeToOpponent, opponentHasBall  \\
\noindent impactingOpponent $\leftarrow$ haveBall, closeToGoal, closeToOpponent \\
\hrule\smallskip
\normalsize
Which, in turn, is embedded in the following ANN:\\
\begin{figure*}[ht!]
	\begin{center}
		\includegraphics[scale=0.25]{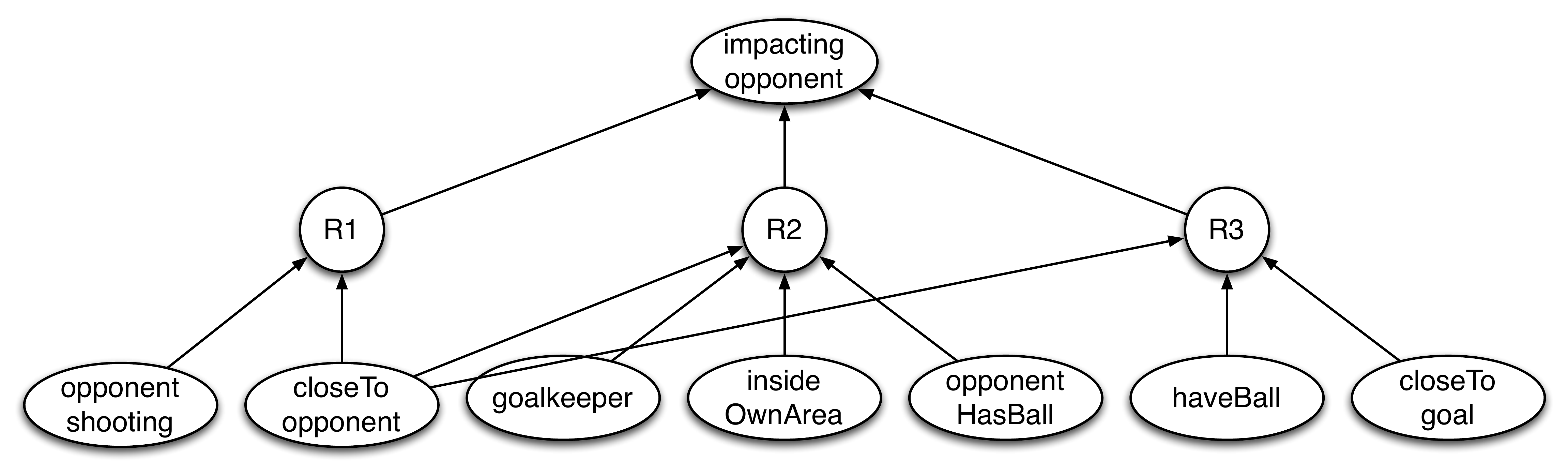}
	\end{center}
\end{figure*}


\subsubsection{Initial Experimental Evaluation of the N-CILP Algorithm}

In order to gain a first idea of the performance and properties of the proposed N-CILP algorithm and the resulting networks, it has been implemented in a proof-of-concept simulator then applied to the above RoboCup example scenario. While the results reported here are still preliminary, they indicate the capabilities of the neural-symbolic approach to normative reasoning and learning under uncertainty.

In the simulator, the KB contains the normative rules that an agent knows. We assume that the priorities are embedded in the rules. The KB is then read as input to the N-CILP translation algorithm, which produces a standard ANN trainable with backpropagation (cf., e.g., \cite{haykin99a}).
The results of training the ANNs are evaluated in the usual way, whereby the performance of a network with random weights initially, i.e. without KB, is compared with that of a network set-up using N-CILP, that is, with KB. Both networks are trained on the same set of examples: pairs of input vectors (opponentShooting, closeToOpponent, etc.) and target output vectors (ImpactingOpponent) with values 1, 0 and -1 denoting, respectively, true, unknown and false. The networks are trained and tested using cross-validation, where the set of examples is divided systematically into a training and a test set, multiple networks are trained and tested on each division (with the test set never seen by the netowork during training), and results are averaged out to produce a better estimate of the network's ability to generalise to new data, that is, its test set performance.

In evaluating the test set performance of the network, two distinct measures are used: \emph{tot} and \emph{part}.
$$tot = \frac{\sum_{i=1}^{n} I(\bigwedge_{j=1}^{k} (c_{ij} == o_{ij})) } {n}$$
$$part = \frac{\sum_{i=1}^{n}\sum_{j=1}^{k} I(c_{ij} == o_{ij})  }{n*k}$$
\noindent Here, $n$ refers to the cardinality of the test set, $k$ is the number of output neurones in the network, $o_{ij}$ is the value of the $j$-th output of the network for the $i$-th test instance, $c_{ij}$ is the target (desired) value of the $j$-th literal for the $i$-th test instance, $I(\cdot)$ is the indicator (i.e. a function returning 1 if the argument is true, and zero otherwise). The {\em tot} measure evaluates how many examples were estimated by the ANN correctly in their entirity (that is w.r.t. the entire target output vector), while \emph{part} measures the average number of output neurones correctly evaluated by the ANN. 

{\bf Comparison with a purely connectionist approach:} The test-set performance of a network built using N-CILP is compared with that of a non-symbolic ANN. One of the well known issues in neural-network training is how to decide the number of neurones in the hidden layer. In the case of N-CILP, this number is given by the number of symbolic rules. We adopt the same number of hidden neurones for both networks and do not perform model selection. The difference between the networks is in the values of the connection weights only. As mentioned, the ANN built with N-CILP sets its weights according to the rules in the KB, whilst the non-symbolic network has its weights initialised randomly. The expected advantage of the network built with N-CILP is that, even without any training, it should be capable of estimating correctly the output value of some of the examples by applying the rules contained in the KB (if the translation is correct, as proved, and the KB is relevant to the data classification problem at hand).

The network built with N-CILP, thus, has the \emph{head-start} of a KB containing rules similar to (and including) the ones used in the example given at the end of the previous section. During the training phase, the network tries to learn additional rules  provided in the form of training examples (input-output vectors). In the interest of fairness, the non-symbolic network is also provided with training examples derived from the initial rules,\footnote{Given a rule, e.g. $B \leftarrow A$, input and output vectors are created having `1' in the position corresponding to $A$ in the input vector, and `1' in the position corresponding to $B$ in the output vector.} but has to learn all rules from scratch using backpropagation. The entire set of rules and preference relations used in our experiments, now with multiple outputs, is given below.\\
\linebreak
\noindent
R1 = (kickoff , O(-score))\\
R2 = (kickoff \& MateTouchesBall , P(score))\\
R3 = (kickoff \& MinBallMoved , P(score))\\
R4 = (True , O(-useHands))\\
R5 = (goalkeeper \& InsideOwnArea , P(useHands))\\
R6 = (True , O(-contactingOpponent))\\
R7 = (True , O(-impactingOpponent))\\
R8 = (impactingOpponent , O(minimizeImpact))\\
R9 = (contactingOpponent , O(terminateContact))\\
R10 = (mateInsideOwnArea , O(-insideOwnArea))\\
R11 = (mateInsideOpponentArea , O(-insideOpponentArea))\\
R12 = (opponentFreeKick , O(keepDistance))\\
R13 = (goalkeeper \& OpponentPenaltyKick \&  -ballTouched , O(-getBall))\\
R14 = (haveBall \& OpponentApproaching , O(pass))\\
R15 = (haveBall \& OpponentApproaching \& OpponentCloseToMate , O(-pass))\\
R16 = (haveBall \& CloseToGoal , O(shoot))\\
R17 = (opponentShooting \& CloseToOpponent , O(impactingOpponent))\\
R18 = (goalkeeper \& InsideOwnArea \& CloseToOpponent \& OpponentHasBall , O(impactingOpponent))\\
R19 = (-goalkeeper \& MateInsideOwnArea \& OpponentShooting , O(-impactingOpponent))\\
R20 = (haveBall \& CloseToGoal \& CloseToOpponent , O(impactingOpponent))\\
R21 = (opponentHasBall \& CloseToOpponent \& CloseToGoal , O(-impactingOpponent))\\
R22 = (-mateInsideOwnArea \& CloseToOpponent \& OpponentHasBall , O(useHands))\\
R23 = (insideOwnArea \& MateInsideOwnArea \& OpponentApproaching , O(-impactingOpponent))\\
R24 = (insideOwnArea \& HaveBall , O(pass))\\
R25 = (opponentFreeKick , O(-canScore))\\
R26 = (opponentPenaltyKick , O(keepDistance))\\     
\linebreak
\noindent
R2 $\succ$ R1 \\  
R3 $\succ$ R1 \\  
R5 $\succ$ R4 \\  
R8 $\succ$ R7 \\  
R9 $\succ$ R6 \\  
R15 $\succ$ R14 \\  
R17 $\succ$ R7 \\  
R18 $\succ$ R7 \\  
R19 $\succ$ R17\\  

\begin{figure}[t]
	\centering
	\includegraphics[width=0.75\linewidth]{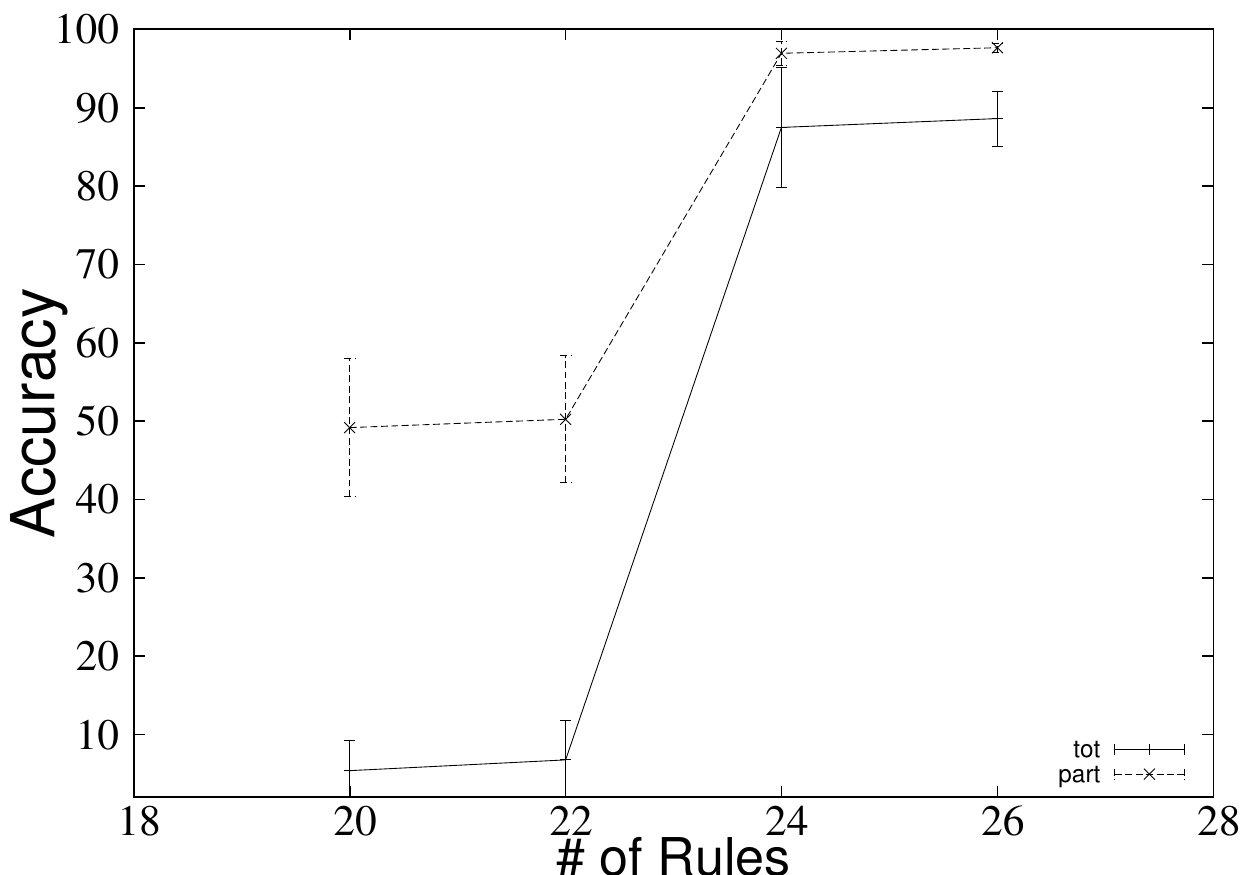}
	\caption{Accuracy of {\em tot} and {\em part} measures for increasing numbers of rules in the knowledge base.
		\label{FIGURA}}
\end{figure}

The results show that the non-symbolic ANN is not able to achieve the same level of accuracy as the N-CILP network. Using the first 20 rules above (R1 to R20) to set up the ANN with N-CILP and the remaining 6 rules (R21 to R26) for testing produced test-set performances $tot = 5.38\%$ and $part = 49.19\%$, while the non-symbolic network achieved $tot = 5.13\%$ and $part = 45.25\%$. More importantly, when we evaluate how the N-CILP ANN perform with increasing number of rules in the KB, test-set performances also increase in a consistent way (see Figure~\ref{FIGURA}. This confirms empirically that the ANN is capable of computing the same semantics as given by the rules in the KB (rules R23 and R24 seem to be particularly relevant), and to exploit learning from examples, which allows a normative agent to increase and adjust its knowledge in the face of multiple possible obligations which may change dynamically in time.

The test is done incrementally using the same 26 rules. The experiment's first run starts with a KB containing the first 20 rules, as before. Subsequently, two additional rules are added to the KB, with each consecutive run decreasing the number of unknown rules that the network has to learn by two, as shown in Figure~\ref{FIGURA}. In the last experiment, with 26 rules, the figure reports the network's traning set performance since there are no rules left from which to derive test set patterns. 

For the first two experiments, accuracy remains low, while for the last two, performance increases considerably reaching a peak of 98,01\% for the \emph{part} measure and 91,18\% for \emph{tot}.

{\bf Learning CTDs:} In a final experiment, we measure the capacity of an ANN built with N-CILP to learn new CTDs. This is done by using a KB with the priority-based orderings that regulate the CTDs left out.

We tested the network on learning three different CTDs, again in the robot-soccer context. The first  refers to a situation where a robot player should never impact on an opponent (R7), but if a collision route is inevitable, then the robot should make its best to minimise the impact (see R7c below). The second CTD addresses a situation where the soccer robot is in physical contact with an opponent, which for most situations is forbidden by standard soccer rules (R6), and should try to terminate the contact (see R6c below). The third CTD handles a situation where, although generally not being allowed to use its hands (R4), the robot finds itself in the role of the goalkeeper (see R4c below). Rules R4, R6 and R7 are reproduced below for convenience.\\
\hrule\smallskip
\small
\noindent R7 = ($\top$ , {\bf O}($\neg$ impactingOpponent))\\   
\noindent R7c = (impactingOpponent , {\bf O}(minimizeImpact)) \\
\noindent R6 = ($\top$ , {\bf O}($\neg$contactingOpponent))\\
\noindent R6c = (contactingOpponent , {\bf O}(terminateContact))\\
\noindent R4 = ($\top$ , {\bf O}($\neg$useHands))\\
\noindent R4c = (goalkeeper \& InsideOwnArea , {\bf P}(useHands))\\
\hrule\smallskip
\normalsize 
\vspace{0.5cm}

Removing the priority-based orderings results in an incomplete system that produces, in similar situations, both the unfulfillable obligation and the relative obligation to handle the suboptimal situation that is being analysed. 
Delivering on the promise to be able to deal with this type of uncertainty in the context of norms, what we expect from our approach is the ability to learn the priority-based orderings that regulate the CTDs. 
The ANN is trained with a set of examples containing both regular situations (R4, R6, R7) and situations in which the CTD is applied (R4c, R6c, R7c). The resulting network is tested with a test set containing situations where an application of the CTD becomes necessary.

For the first CTD, results show a $95\%$ test-set performance by the network, which generated \emph{minimizeImpact} only when in the suboptimal CTD situation in question. For the two other CTDs, the results show an accuracy of $93\%$ and $87\%$ on their respective test sets.
This indicates that N-CILP is capable of learning CTDs not included in the construction of the ANN. It, thus, allows us to avoid a total description of the corresponding domain (which very often turns out overly expensive or simply infeasible) as missing norms can be acquired through learning from examples.

\section{Conclusion}\label{conclusion}

At the beginning of Section~\ref{introduction} we set out to argue two connected claims. Firstly, we aimed to show that probability is not the only way of dealing with uncertainty (and even more, that there are kinds of uncertainty which are for principled reasons not addressable with probabilistic means). Secondly, we wanted to provide evidence that logic-based methods can well support reasoning with uncertainty, using two paradigmatic examples: LP with Kleene semantics for modelling reasoning from information in a discourse, to an interpretation of the state of affairs of the intended model, and a neural-symbolic implementation of a fragment of I/O logic expressed as extended LP for dealing with uncertainty in dynamic normative contexts. Looking back at what has been reported in the previous sections, we believe that both goals have been met. Even more, while at first sight seeming fairly independent from each other, we hope that also the intrinsic---formal and conceptual---connection between LP for reasoning to an interpretation on the one hand, and the neural-symbolic I/O logic approach combining normative reasoning and learning on the other hand, have become apparent. The neural-symbolic I/O setting presents a natural expansion of the LP approach. In addition, the normative features also (via the additional ANN characteristics) add learning capacities to the previously exclusively reasoning-focused framework.\footnote{The presented approach to LP modelling of discourse does not tackle the learning of KB rules, as discourse comprehension generally is assumed to proceed with a mature KB. But an account of learning is nevertheless an important goal for LP models of discourse.} Still, it should be clear that the discussed account of LP and neural-symbolic I/O logic are only two examples among several for logic-based methods dealing with forms of uncertainty, and that even for these two the presented work can only be considered initial steps in the direction of fully exploring---and exploiting---the possibilities offered by the respective approaches beyond the use of probabilistic models. 

As a general insight gained from our described explorations into uncertainty and logical methods, we note that in fact examining the nature of the uncertainty and its twinned necessity in each logic provides at least a semi-systematic method of exploring for species of uncertainty. As we noted, at least LP, and deontic logics provide examples which are clearly interesting for human cognition and the modelling thereof with computational means. These are the first which we have examined in any detail. We do not claim that every logic has its own distinct species, nor that every species enumerated in this way is of any interest to cognitive modelling or AI. However, even from these examples, it is clear that logic can serve as a royal road to the exploration (and handling) of different kinds of uncertainty. The only generalisation we would offer at this point is that logics differ in their kind of uncertainty insofar as they specify distinct kinds of epistemic state. It is the epistemic states that cannot always be matched by other logics that give rise to different kinds of uncertainty, rather than some general property of the inferences that are valid, or the content of their propositions.

Concerning future work, it seems desirable to also develop an architecture combining the described form of LP modelling with neural-symbolic computing analogous to the I/O logic setting. As discussed, for instance, by \cite{stvl06:book}, there is already a neural implementation for simple LP.  Constraint LP is a more expressive logic which includes the Event Calculus, and is required for modelling, among other things, all but the simplest reasoning in the processing of time and causality in narrative discourse \citep{laha04:prop}. A neural network implementation for this formalism is currently lacking, but would most likely have great advantages: on the one hand, introducing the ANN characteristics as part of the neural-symbolic implementation would allow the introduction of  learning capacities into the discourse processing context, expanding the approach and corresponding model in a natural way. On the other hand the availability of such an architecture would further bridge from the currently still (mostly) cognitive modelling-oriented setup to applications of the paradigm in corresponding models in cognitively-inspired AI.

Regarding the neural-symbolic implementation of I/O logic, we next hope to introduce an explicit notion of context in the neural-symbolic system. In reality, choices are not made by only taking into consideration the current situation, but are usually also influenced by past events. Continuing with the robot soccer example, for instance we might want to consider situations where a robot changes its style of play due to the previous and current history: yellow cards received would make the robot play in a safer way to avoid being sent off; if the current result suggests that the robot team is already winning, they could prefer to play  more defensively to prevent the other team equalising. In order to implement those mechanisms, the system must be capable of memorising past events. One way to solve this might be to add external memory to the networks \citep{bordes}. With this solution the context nodes in the ANN could be added in the same way as for the rules, the difference being that for each context in the input level there would be a correspondent output context which is linked from the output to the input levels, in order to maintain memory if any context modified its status during computation. A related line of potential future research involves the area of argumentation. Argumentation has been proposed, among other things, as a method to help symbolic machine learning. In \cite{mozina}'s approach, an expert's reasons for some of the training examples can be used to try and guide the search for hypotheses, in a way similar to our use of background knowledge \citep{JLC05arguma}.

In a third line of development on the systems-oriented side, we want to take the neural-symbolic architecture for I/O and---reusing insights from the LP approach described in this article--- develop a follow-up framework additionally modelling interpretation-related aspects of reasoning. If successful, this would allow to address the case when one does not know which propositions are actually relevant (i.e. combining reasoning to an interpretation with subsequent normative reasoning while maintaining the ability to deal with dynamically changing sets of rules).

From a conceptual perspective, we would like to get clearer how uses of intensional and extensional systems---as already discussed in Section~\ref{introduction}---might work together. \cite{kslm15:decision} argue that systems that use extension sets to capture the meanings of predicates---at least when those systems are used for cognitive modelling---necessarily rely for their foundations on intensional systems that can capture the interplay of  motivations of the reasoner (desires, purposes, goals, preferences, \ldots).  The extensional systems `precisify' or perhaps operationalise intensional meanings in specific contexts.  But different extensional precisifications of the same  intensional concept may be incompatible in having different extensions. Intensional systems can capture the crucial abstractions due to the flexibility of motivational elements, answering the question `Why this extension in this interpretation?'. Extensional systems are important, but their importance cannot be understood without understanding their basis in intensional systems.  The issues of operationalising concepts for statistical modelling are commonplace to psychologists, but analogous decisions have to be made in many other related domains, including everyday discourse. If we are reasoning about the reliability of the conditional ``If the brake pedal is pressed, the car slows down.'' then the extension of cars excludes ones on the dump. If a mechanic is searching for a spare part, and reasons about the conditional: ``If the car is a 2009 or later, it complies with the emissions regulations.'' then the ones on the dump may be exactly the ones that are in the relevant extension. We negotiate extensions for `car' through our intensional purposes for reasoning, and when we construct them, they do not replace the vague intensional meanings that went into their construction. 
Reiterating a point already argued in the introduction,  ``intensional'' systems like LP with Kleene semantics can express goals in a sense which is not fully possible in ``extensional'' systems like probability theory. So, once again, it is  important to distinguish the different kinds of uncertainty they treat. At the most general level, this paper is an argument for a strategy in understanding uncertainty.  The novel kinds of uncertainty exemplified here are of a rather extreme kind. Establishing extreme examples is important.  Extreme examples may not make good law, but they greatly aid exploration.    

\section*{Acknowledgements}
	We want to thank the following people for their indispensable contributions to different parts of the work reported in this article: Guido Boella, Silvano Colombo Tosatto, Valerio Genovese, Laura Martignon, Alan Perotti, and Alexandra Varga.

\bibliographystyle{plainnat}
\bibliography{vagueness_r2_arxiv}   

\end{document}